\documentclass[preprint,authoryear]{elsarticle}

\usepackage{natbib}
\usepackage{lineno}
\usepackage{amsmath}
\usepackage{amssymb}
\usepackage{graphicx}
\usepackage{subcaption}
\usepackage{url}
\usepackage{hyperref}
\hypersetup{hidelinks}
\usepackage{siunitx}
\usepackage{booktabs}
\usepackage{multirow}
\usepackage{array}
\usepackage{colortbl}
\newcolumntype{L}[1]{>{\raggedright\let\newline\\\arraybackslash\hspace{0pt}}m{#1}}
\newcolumntype{C}[1]{>{\centering\let\newline\\\arraybackslash\hspace{0pt}}m{#1}}
\newcolumntype{R}[1]{>{\raggedleft\let\newline\\\arraybackslash\hspace{0pt}}m{#1}}
\usepackage[linesnumbered,ruled,vlined]{algorithm2e}
\usepackage[section]{placeins}
\usepackage{rotating}
\usepackage{tikz}
\usepackage{pgfplots}
\pgfplotsset{compat=1.18}
\usepgfplotslibrary{groupplots,fillbetween}
\usetikzlibrary{shapes, shapes.geometric, arrows.meta, positioning, fit, calc, backgrounds}

\DeclareMathOperator*{\argmax}{arg\,max}

\definecolor{revcolor}{RGB}{176,0,110}
\definecolor{bestcell}{rgb}{0.84,0.94,0.84}
\definecolor{ppocell}{rgb}{0.89,0.83,0.94}
\definecolor{neatcell}{rgb}{0.99,0.89,0.75}

\journal{Transportation Research Part C}

\begin{document}

\begin{frontmatter}

\title{Emergent Charging Coordination in Electric Delivery Fleets}

\author[upct]{Javier Vales-Alonso\corref{cor1}}
\ead{javier.vales@upct.es}
\cortext[cor1]{Corresponding author.}
\author[upct]{Juan J. Alcaraz}
\ead{juan.alcaraz@upct.es}
\affiliation[upct]{organization={Department of Information and Communication Technologies, Universidad Polit\'ecnica de Cartagena},
            city={Cartagena},
            country={Spain}}


\begin{abstract}
In electric delivery fleets, mid-shift charging is non-trivial: each
vehicle must decide when, where and how much to charge to finish on time
with battery above a safety floor.  The choices are coupled: queues
build where too many vehicles pick the same station.  Prior work
resolves this coupling with central dispatching, precomputed schedules
or reservations, machinery that charging infrastructure rarely
supports.  Instead, we use a family of learning agents
under purely local control: every vehicle runs the same policy, deciding
alone from its time budgets and broadcast station occupancies, leading
to emergent coordination without central control or messaging.
We validate this paradigm in simulation on real OpenStreetMap networks of twenty
cities, each with a frozen scenario calibrated by an omniscient Oracle
(99.5\% of shifts completed on time), whereas a
naive greedy rule (nearest station on low battery) completes just 73\%.  Agents
trained with neuroevolution (NEAT) and policy gradients (PPO) on four
cities and deployed zero-shot across all twenty, sixteen never seen in training,
complete 96.8\% and 98.6\% of shifts, with the policy-gradient controllers proving
more robust when demand or vehicle characteristics drift beyond the trained regime.  In contrast, tuned threshold heuristics
that read vehicle urgency alone fall short in
contended cities ($\sim$80\%).
Through training, these learning agents rediscover partial charging and short
opportunistic sessions, and route around busy stations, cutting per-session queue
waits from about 45 minutes to under 2.
In summary, this coordination paradigm balances local
urgency against public occupancy, reaching near-Oracle performance at
minimal implementation cost.
\end{abstract}


\end{frontmatter}

\section{Introduction}
\label{sec:intro}

\begin{figure}[t]
\centering
\begin{tikzpicture}[scale=0.75, every node/.style={transform shape},
    >=Stealth,
]

\definecolor{soc80}{HTML}{2E7D32}   
\definecolor{soc65}{HTML}{43A047}   
\definecolor{soc50}{HTML}{9CCC65}   
\definecolor{soc35}{HTML}{FFA000}   
\definecolor{soc20}{HTML}{FF6F00}   
\definecolor{soc10}{HTML}{E53935}   
\definecolor{routeblue}{HTML}{1565C0}
\definecolor{chargeorange}{HTML}{E67E22}

\newcommand{\van}[3]{
    \begin{scope}[shift={(#1,#2)}, scale=#3]
        \fill[blue!18!white, rounded corners=0.8pt] (-0.8,-0.35) rectangle (0.5, 0.35);
        \fill[blue!28!white, rounded corners=0.8pt] (0.5,-0.35) -- (0.5,0.35) -- (0.9,0.35) -- (1.1,0.0) -- (1.1,-0.35) -- cycle;
        \fill[white, opacity=0.7] (0.55,0.0) -- (0.55,0.28) -- (0.82,0.28) -- (0.95,0.0) -- cycle;
        \fill[black!65] (-0.45,-0.35) circle (0.12);
        \fill[black!65] (0.75,-0.35) circle (0.12);
        \draw[black!50, line width=0.4pt, rounded corners=0.8pt] (-0.8,-0.35) rectangle (0.5, 0.35);
        \draw[black!50, line width=0.4pt] (0.5,-0.35) -- (0.5,0.35) -- (0.9,0.35) -- (1.1,0.0) -- (1.1,-0.35) -- cycle;
    \end{scope}
}

\newcommand{\battbar}[5]{
    \fill[black!6, rounded corners=0.5pt] (#1,#2) rectangle ++(#3, 0.20);
    \fill[#5] (#1,#2) rectangle ++(#3*#4, 0.20);
    \draw[black!35, line width=0.4pt, rounded corners=0.5pt] (#1,#2) rectangle ++(#3, 0.20);
}

\newcommand{\csicon}[3]{
    \begin{scope}[shift={(#1,#2)}, scale=#3]
        \fill[black!8, rounded corners=1pt] (-0.45,-0.5) rectangle (0.45, 0.5);
        \draw[black!45, line width=0.6pt, rounded corners=1pt] (-0.45,-0.5) rectangle (0.45, 0.5);
        \fill[green!20] (-0.25, 0.05) rectangle (0.25, 0.35);
        \draw[black!35, line width=0.3pt] (-0.25, 0.05) rectangle (0.25, 0.35);
        \fill[black!50] (0.02,0.32) -- (-0.08,0.18) -- (0.0,0.18) -- (-0.02,0.08) -- (0.08,0.22) -- (0.0,0.22) -- cycle;
    \end{scope}
}

\newcommand{\conggauge}[5]{
    \fill[black!4, rounded corners=0.5pt] (#1,#2) rectangle ++(#3, 0.22);
    \fill[#5] (#1,#2) rectangle ++(#3*#4, 0.22);
    \draw[black!25, line width=0.3pt, rounded corners=0.5pt] (#1,#2) rectangle ++(#3, 0.22);
}


\csicon{0.8}{2.8}{1.0}
\conggauge{0.15}{1.85}{1.3}{0.5}{yellow!60!orange}
\node[font=\footnotesize\bfseries, black!60] at (0.8, 1.55) {CS\textsubscript{1}};

\csicon{4.0}{0.3}{1.0}
\conggauge{3.35}{-0.55}{1.3}{0.15}{green!50}
\node[font=\footnotesize\bfseries, black!60] at (4.0, -0.85) {CS\textsubscript{2}};

\csicon{7.5}{2.8}{1.0}
\conggauge{6.85}{1.85}{1.3}{0.85}{red!50}
\node[font=\footnotesize\bfseries, black!60] at (7.5, 1.55) {CS\textsubscript{3}};

\node[rectangle, fill=black!75, draw=black!90, minimum size=4mm, inner sep=0pt, line width=0.8pt] (depot) at (0.8, 4.2) {};
\node[font=\footnotesize\bfseries, black!60, left=3pt] at (depot) {Depot};

\node[circle, fill=routeblue!25, draw=black!40, minimum size=4.5mm, inner sep=0pt, line width=0.8pt] (p1) at (2.5, 4.8) {};
\node[circle, fill=routeblue!25, draw=black!40, minimum size=4.5mm, inner sep=0pt, line width=0.8pt] (p2) at (4.2, 5.8) {};
\node[circle, fill=routeblue!25, draw=black!40, minimum size=4.5mm, inner sep=0pt, line width=0.8pt] (p3) at (5.8, 5.0) {};

\node[circle, fill=routeblue!10, draw=black!20, minimum size=4.5mm, inner sep=0pt, line width=0.6pt] (p4) at (6.8, 3.8) {};
\node[circle, fill=routeblue!10, draw=black!20, minimum size=4.5mm, inner sep=0pt, line width=0.6pt] (p5) at (7.8, 5.2) {};
\node[circle, fill=routeblue!10, draw=black!20, minimum size=4.5mm, inner sep=0pt, line width=0.6pt] (p6) at (6.2, 6.6) {};
\node[circle, fill=routeblue!10, draw=black!20, minimum size=4.5mm, inner sep=0pt, line width=0.6pt] (p7) at (3.5, 6.6) {};

\draw[routeblue, line width=1.8pt, ->] (depot) -- (p1);
\draw[routeblue, line width=1.8pt, ->] (p1) -- (p2);
\draw[routeblue, line width=1.8pt, ->] (p2) -- (p3);

\draw[chargeorange, line width=1.6pt, densely dashed, ->] (p3) -- (4.0, 1.0);
\draw[chargeorange, line width=1.6pt, densely dashed, ->] (4.0, 1.0) -- (p4);

\draw[black!15, line width=0.7pt, dashed, ->] (p3) -- (p4);

\draw[routeblue!40, line width=1.2pt, dashed, ->] (p4) -- (p5);
\draw[routeblue!40, line width=1.2pt, dashed, ->] (p5) -- (p6);
\draw[routeblue!40, line width=1.2pt, dashed, ->] (p6) -- (p7);
\draw[routeblue!40, line width=1.2pt, dashed, ->] (p7) -- (depot);

\van{5.8}{5.65}{0.38}
\battbar{5.42}{5.92}{0.80}{0.18}{soc20}

\van{5.2}{1.9}{0.38}
\battbar{4.82}{2.17}{0.80}{0.70}{soc65}

\end{tikzpicture}
\caption{Station-selection tradeoff. A van at stop~3 with low SoC (orange node) detours to lightly loaded CS\textsubscript{2} instead of congested CS\textsubscript{3}, exits at partial charge, and resumes the route. Node colors encode SoC (green\,=\,high, red\,=\,low); faded nodes are future stops.}
\label{fig:system}
\end{figure}
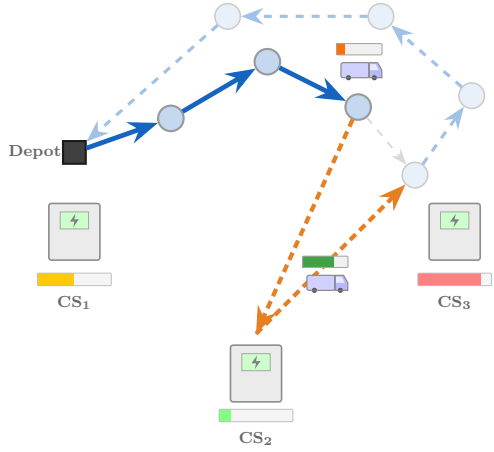

\begin{figure*}[t]
\centering
\includegraphics[width=\textwidth]{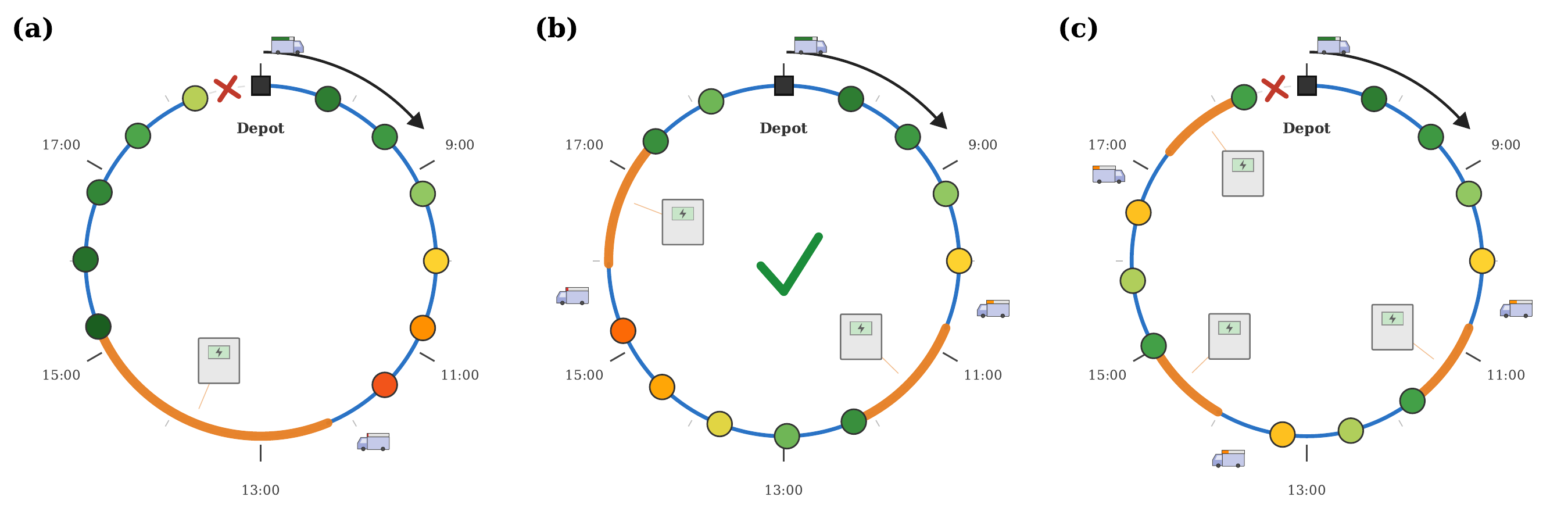}
\caption{Three charging strategies on a 12-stop route, where angular position encodes elapsed time within a 12-hour shift (clockwise from 7:00). Stop circles are colored by SoC (green\,=\,high, red\,=\,low); orange arcs mark time spent at charging stations. (a)~Single charge to 100\,\%: the slow CV phase above 80\,\% SoC turns a single stop into a 180-minute block, causing the shift to expire before the last stop is reached. (b)~Two partial charges to $\sim$70\,\%: staying in the fast CC region keeps each stop under 90\,min, and the route completes with margin. (c)~Three shorter charges to $\sim$65\,\%: each stop is only 60\,min, but the cumulative overhead of three detours again exceeds the shift budget.}
\label{fig:charging_strategies}
\end{figure*}

Delivery fleets are shifting from fossil fuels to electric power,
driven by emission regulations and falling battery costs~\citep{IEA2025gevo}.
This transition introduces operational constraints that conventional vehicles
never faced.  Refueling a fossil-fuel vehicle takes minutes at virtually any
service station; an Electric Vehicle (EV), by contrast, carries less range per
full charge, requires specific infrastructure, Electric Vehicle Supply
Equipment (EVSE) connectors, and recharges far more slowly, especially above
80\% state-of-charge (SoC), where charging power tapers off.  During a working
shift, an EV on a longer route, with a smaller battery, or starting only partly
charged often cannot finish on a single charge and must stop to charge, one or more
times, at charging stations potentially shared with other users.  Not every
fleet faces this: on light last-mile routes under about $100$~km a day, charging
overnight at the depot is usually enough~\citep{Pelletier2018}.  Metropolitan
delivery is different, and there a significant share of the fleet must top up
mid-shift~\citep{Yang2024urbanev}.
In the settings we study, about half of the vehicles cannot finish a
shift on their starting charge.  Even fleets of larger-battery
vehicles that begin a shift nearly full are not exempt: fewer need to charge, but
the number of EVs per charging point keeps rising worldwide~\citep{IEA2025gevo},
so those that do face sharper competition for the same chargers.
Because each station has only a few
connectors, a vehicle that finds them all busy must queue or drive on to
another~\citep{Kullman2021,Froger2022}.  Coordinating this shared charging is the
problem we address.
Each charging visit forces two trade-offs.  The first is \emph{where}
to charge: the nearest station is not always the best choice, and driving past a
congested station to reach a more distant but lightly loaded one can save time
overall (Fig.~\ref{fig:system}).  The second is \emph{how much} to
charge: a full charge keeps the vehicle out of service far longer and can
exhaust the shift, whereas a quick partial charge returns it to the route sooner
but could force another stop later, with no guarantee that a connector will then be
free (Fig.~\ref{fig:charging_strategies}).

Therefore, each vehicle that needs to charge faces three coupled decisions:
\emph{whether} to charge now, \emph{at which station}, and \emph{to what SoC}.  Every choice constrains the ones that
follow, and---because stations are shared---the choices of one vehicle can constrain
those of the rest.  All of this must be decided under time pressure, at every
delivery stop, by every vehicle in the fleet.  The literature typically
addresses these decisions in isolation (Section~\ref{sec:related}); even the
closest approaches resolve at most two of the three, relying on centralized
dispatching with full fleet telemetry~\citep{Tuchnitz2021smart},
centrally trained multi-agent control over a graph of the whole fleet~\citep{Zhou2022gmix}, or single-vehicle dynamic routing with
station occupancy data~\citep{Dastpak2024trc}.  Such requirements are rarely met
in practice.  The most that a shared station can realistically be
expected to expose is a real-time congestion signal---how many connectors are
busy, with no visibility of the queue behind them---readily available through
the Open Charge Point Protocol (OCPP) and requiring no cooperation from the
operator beyond basic status data.

This gap between what the literature requires and what the infrastructure
offers motivates the central idea of this paper: \emph{emergent coordination}.
Each vehicle decides its charging actions independently, using only its own
state---battery level, route progress, and time left in the shift, which
together signal its \emph{charging urgency}---and the publicly broadcast
congestion at nearby stations.  No reservations are made and no messages are
exchanged, neither with other vehicles nor with any control center, and no
central entity coordinates the fleet.  Coordination is not designed in; it
\emph{emerges} through training, as every vehicle runs the same learned
controller and learns to weigh the broadcast congestion against its own urgency.
The same principle already works in large-scale EV charging, where decentralized
agents responding to a common price signal converge on a stable coordinated
outcome without central control~\citep{Ghavami2024pricing}.  We extend it to the
full charging problem---all three decisions resolved at once---for metropolitan
delivery fleets on real road networks.

This coordination is a property of the \emph{interface}---the signals each
vehicle reads and the rule mapping them to a charging action---not of any
particular learning method, and we show it by instantiating that interface in
different ways.  The controllers come from two
families that are among the strongest neural-policy learners in use today:
neuroevolution (NeuroEvolution of Augmenting Topologies,
NEAT)~\citep{Stanley2002neat} and policy-gradient reinforcement learning
(Proximal Policy Optimization, PPO)~\citep{Schulman2017ppo}.  Each family is
realized in two architectures: a \emph{per-candidate} controller that scores
each candidate station independently, and a \emph{joint} controller that
processes all candidate inputs at once.

Controllers are trained and evaluated in a realistic discrete-event simulator
built on real cities: road graphs, charging stations and delivery Points of
Interest (POIs) drawn from OpenStreetMap via OSMnx~\citep{Boeing2017osmnx}, using a per-city calibrated statistical
routing surrogate for distances and travel times.  The simulator also captures the physics of EV
charging---constant-current/constant-voltage (CC/CV) charging with a continuous
taper above 80\% SoC, heterogeneous station power and per-vehicle limits---together
with the congestion of stations shared with other users.
Controllers have been trained (${\sim}300$ CPU-hours each) on scenarios from
four real cities (Madrid, Los Angeles, Chongqing, S\~ao Paulo) and benchmarked on
16 more across five continents (Paris, Tokyo and New York among them).  We
measure them against an omniscient Oracle: a planner that knows each
shift's entire future.

Among the main findings, the strongest generalist controller (PPO-joint) completes 98.6\% of
delivery shifts on time across the 20-city benchmark zero-shot, statistically close to the
omniscient Oracle reference (99.5\%).  Neuroevolution proved a fast, gradient-free way to
validate the paradigm, while the policy-gradient controllers reached a higher level at the
cost of a markedly more complex training process.  The learned policies are also interpretable: built on a handful of decision signals, they let
us read why each charging choice is made, and they rediscover strategies the reward never
encoded, such as partial charging and routing around congested stations.

The remainder of this paper is organized as follows.  Section~\ref{sec:related} reviews related work.  Section~\ref{sec:problem} states the operational problem.  Section~\ref{sec:agent} presents the agent design, and Section~\ref{sec:learning} the learning algorithms and baselines that realize it.  Section~\ref{sec:system} describes the evaluation environment, and Section~\ref{sec:experiments} the setup and benchmark, the training strategy, and the results.  Section~\ref{sec:conclusions} concludes the paper.

\section{Related Work}
\label{sec:related}

Fleet charging has drawn sustained academic attention, with proposals that
differ in their assumptions, principles and mechanisms
(Table~\ref{tab:comparison}).  This section reviews the lines of work most
closely related to ours.  For a broad overview of the problems and results in
the area we refer the reader to recent surveys~\citep{Zhao2024tre,
Kucukoglu2021survey} and references therein.
Two axes separate these lines of work.  The first is when the charging decision
is committed: ahead of the shift (offline) or as it unfolds (online).  The second
is where the deciding rule comes from: from solving or designing a model, or from
learning it against a reward.

\begin{table*}[t]
\centering
\begin{scriptsize}
\setlength{\tabcolsep}{3pt}
\begin{tabular}{L{3.1cm} L{2.0cm} L{1.5cm} C{1.0cm} C{1.0cm} C{1.4cm} L{2.5cm} L{2.0cm}}
\toprule
\textbf{Reference} & \textbf{Family} & \textbf{Goal} & \multicolumn{3}{c}{\textbf{Decisions}} & \textbf{Info required} & \textbf{Network} \\
\cmidrule(lr){4-6}
                   &                 &               & \textit{When} & \textit{Where} & \textit{How-much} & & \\
\midrule
\citet{Bragin2024trc}       & Offline $\cdot$ opt   & Cost           & \checkmark & \checkmark & \checkmark & Central telem. & Real \\
\midrule
\citet{Xu2016priority}      & Online $\cdot$ opt    & Cost           & \checkmark & --         & \checkmark & Central telem. & --   \\
\citet{Iacobucci2019mpc}    &                  & Cost + wait    & \checkmark & \checkmark & \checkmark & Central telem. & Real \\
\citet{Mahyari2025adp}      &                  & Cost           & \checkmark & --         & \checkmark & Central telem. & Depot \\
\midrule
\citet{Tuchnitz2021smart}   & Online $\cdot$ learn  & Grid load      & \checkmark & --         & \checkmark & Central telem. & Resid. \\
\citet{Shi2020fleet}        &                  & Cost + wait    & \checkmark & --         & --         & Central telem. & Synth. \\
\citet{Liao2025trc}         &                  & Travel time    & --         & \checkmark & --         & Central telem. & Real \\
\midrule
\citet{DaSilva2020marl}     & Online $\cdot$ learn (comm.) & Grid load & \checkmark & --         & --         & Communication  & Resid. \\
\citet{Alqahtani2022marl}   &                  & Energy cost    & \checkmark & \checkmark & partial    & Central telem. & 5$\times$4$\to$Synth. \\
\citet{Zhou2022gmix}        &                  & Throughput     & \checkmark & --         & --         & Central telem. & Synth. \\
\citet{Mei2026apen}         &                  & Cost           & \checkmark & \checkmark & \checkmark & Central telem. & Coupled P-T \\
\citet{Ahadi2023coop} &                  & Profit + service & \checkmark & \checkmark & --         & Central telem. & Synth. \\
\midrule
\citet{Dastpak2024trc}      & Online $\cdot$ signal & Travel time    & --         & \checkmark & \checkmark & Public signal   & Synth. \\
\citet{Ghavami2024pricing}  &                  & Congestion     & --         & \checkmark & --         & Public price   & Real \\
\citet{Popiolek2023reservation} &              & Travel time    & \checkmark & \checkmark & --         & Communication  & Real \\
\midrule
\textbf{This work} & \textbf{Online $\cdot$ learn $\cdot$ signal} & \textbf{On-time + SoC floor} & \textbf{\checkmark} & \textbf{\checkmark} & \textbf{\checkmark} & \textbf{Public + local signal} & \textbf{Real (OSMnx)} \\
\bottomrule
\end{tabular}
\caption{Representative approaches for EV fleet charging coordination, grouped by family.  Each \emph{Family} label names a cell of two axes---when the charging decision is committed (offline or online) and where the deciding rule comes from (optimization or learning)---together with the coordination channel the method assumes.  \emph{Goal}: the objective each method optimizes; most target cost, grid load or travel time, and only this work targets the fleet's on-time service level while keeping every battery above its safety floor.  \emph{Decisions}: whether a method resolves \emph{when}, \emph{where} and to \emph{what level} to charge.  \emph{Info required}: the coordination information the method assumes --- central telemetry, inter-agent communication, a reservation layer, or only a broadcast public signal.  Alone in the table, this work resolves all three decisions from a public broadcast signal on a real road network.}
\label{tab:comparison}
\end{scriptsize}
\end{table*}

The oldest and best-developed approach solves everything offline, by optimization.
The most closely related problem is the Electric Vehicle Routing Problem
(EVRP)~\citep{Kucukoglu2021survey}, which plans routes and charging together.
It accommodates features such as heterogeneous recharging technologies, nonlinear
charging functions and time windows, and recent exact methods keep extending
these variants~\citep{Nafstad2025trsc}.  In contrast to the EVRP, our delivery routes are already given, so a charging
stop is a detour on a fixed route, not a routing decision.  We
adopt this assumption because the reality is dynamic (queues, delays and the
other events that matter cannot be known in advance) whereas an EVRP plan
presupposes exactly that knowledge.  In dense urban delivery these uncertainties only sharpen, with unpredictable traffic, sudden street closures and volatile station occupancy.  Offline optimization still teaches one lesson: charging must be designed with
care, since trivial policies can fail to meet operational requirements.
\citet{Bragin2024trc} formulates routing and charging jointly, arguing that
decoupling them is suboptimal, and the same pattern recurs up to network-wide
bus charging schedules~\citep{Zhou2024buscharging}.

Online, non-learned methods come in several forms.  The simplest are fast rules
such as priority orderings and SoC thresholds that decide charging locally and in
constant time~\citep{Xu2016priority}.  We keep one such rule, a hand-designed threshold heuristic, as a non-learned baseline (Section~\ref{sec:sa_heuristic}).  Model predictive control (MPC) adds
look-ahead: \citet{Iacobucci2019mpc} runs two receding-horizon controllers at two
timescales for a shared-autonomous EV fleet on a real case, an approach still
active~\citep{Skugor2025mpc}.  Approximate dynamic programming (ADP) approximates
the optimal policy rather than re-solving at each step~\citep{Mahyari2025adp,
Lee2022adp}.  These methods work well where their assumptions hold, but they sit in a different
regime from ours.  As applied to fleet charging, MPC and ADP are central
optimizers: they work from fleet-wide state and re-plan for the whole fleet.  Our
setting removes that central optimizer---each vehicle decides on its own, from its
state and a broadcast signal.

Like the methods above, reinforcement learning decides online.  But it learns the
rule from reward instead of fixing it, gaining the adaptability that offline plans
lack.  Its record on this problem, though, is fragmented~\citep{Zhao2024tre}.  Timing, station
selection and energy target are usually treated as separate problems, and the
strongest single-agent results resolve only one or two of the three, under
simplified environments.  Each removes the very difficulty we target, the
station choice: charging is pinned to a dedicated home
charger~\citep{Tuchnitz2021smart}, confined to a single
station~\citep{Ye2022cade}, or set on a synthetic zone grid where the vehicle
always charges at its nearest station~\citep{Shi2020fleet}.
These results are also scored on different terms.  The RL-charging literature reports energy and cost proxies (charge variance, electricity cost, queue time, constraint satisfaction) not a fleet service level such as the share of shifts completed on time.  The closest analog, \citet{Tuchnitz2021smart}, reports no energy-shortfall cases and a 65\% reduction in charging variance, but for overnight home charging with one charger per vehicle, without station selection or queues.
The real difference, though, is at run time, in the coordination channel each design assumes.  Distributed multi-agent controllers such
as~\citet{DaSilva2020marl} coordinate by having the vehicles exchange messages.
Ours exchanges nothing, each vehicle acting on its own state and the public
congestion signal alone.  The closest recent work, \citet{Mei2026apen}, resolves routing and charging
jointly for a fleet, with charging coupled to the power grid through price
signals.  In the same spirit, \citet{Ahadi2023coop} show that decentralized,
learned charging can outperform centralized static control for a shared fleet,
though their vehicles still answer to a central coordinator that steers them from the
fleet's average state.

Our work fits the online, learned, signal-only category, and to our knowledge it
is the first of its kind.  The emergent coordination it relies on---coherent
collective behavior arising from purely local reactions to a shared signal, with
no central control---has been investigated in other settings.
The intuition is
classical: selfish agents reacting to congestion settle into predictable
equilibria~\citep{Wardrop1952, Rosenthal1973}.  It has been realized in
decentralized charging, where vehicles respond to a broadcast price with no
central controller dictating their actions~\citep{Ghavami2024pricing}, and in mean-field models of large
EV populations~\citep{Tajeddini2019meanfield}.  As in our system, these agents
react to a broadcast signal alone, with no messages between vehicles.  Each reports consistent gains, from single-vehicle routing to energy markets~\citep{Dastpak2024trc,
Pournaras2018collective, SalazarPena2026marlem, Ge2025distributed}. The conclusion is that a single public signal bearing enough information removes
an entire layer of coordination infrastructure and simplifies system design.

\section{The Operational Problem}
\label{sec:problem}
\label{sec:cccv}

A delivery operator runs a fleet of $N$ EVs out of one or more depots.  Each
vehicle $i$ has a battery of capacity $C_i$ (kWh) and an energy consumption rate
$e_i$ (kWh/km), jointly summarized by the normalized consumption
$\zeta_i = e_i / C_i$, the fraction of battery spent per kilometer.  For each vehicle, a working shift is
characterized by a route (an ordered set of delivery stops), a time
budget $h_i$, an initial state of charge $\text{SoC}_i^0$ and a departure delay
$\delta_i$ from the depot. These change from shift to shift, with departures staggered so the fleet does
not leave all at once.  Each shift is an independent \emph{episode}: fresh routes
and initial charge, with no state carried over from the previous one.  Route and shift are bound together: one determines the
other, so that driving,
service and the return to the depot fit within $h_i$, of which a fraction
$\varphi h_i$ is held back as a cushion to absorb delays, charging sessions, and other
events.  Charging may
happen at any point of the shift, at $M$ \emph{shared} stations, infrastructure
that serves the fleet and, potentially, vehicles outside it.  The
charging physics and the station-congestion dynamics are described next
(Sections~\ref{sec:physics} and~\ref{sec:congestion}). Table~\ref{tab:params}
lists example values used in the evaluation.

A vehicle's route ends in one of three ways: it is completed within the shift
(success), the battery falls below a safety floor $\text{SoC}_{\min}$ (failure),
or the shift runs out before the route is completed and the vehicle is back at the
depot (failure).  The goal is to maximize the success ratio, the fraction of
vehicles that succeed.  We assume each vehicle runs an instance of the same
control agent, deciding independently when, where and how much to charge
(Section~\ref{sec:agent}).  A further design assumption is that a vehicle decides
only at discrete points --- at each route stop, just before departing for the next
--- with no intermediate decisions.

\subsection{Charging physics}
\label{sec:physics}
Each station $j$ has $E_j$ connectors of a single nominal power $P^{\text{evse}}_j$,
identical within the station, as is common in practice.  Stations differ in power
across the network, from AC charging to DC fast charging.  Each vehicle
carries a Battery Management System
(BMS) cap $P^{\text{bms}}_i$ on the power it can accept.  Both the station powers
and the caps vary across scenarios, with typical ranges in Table~\ref{tab:params}.  The effective sustained power for vehicle $i$ at station
$j$ is limited by both ends of the chain:
\begin{equation}
P^{\text{eff}}_{ij} = \min\left(P^{\text{evse}}_j,\; P^{\text{bms}}_i\right),
\label{eq:power}
\end{equation}
with the electrical installation sized to serve all connectors at nominal
power simultaneously.  Charging then follows the
constant-current/constant-voltage (CC/CV) profile of lithium
batteries~\citep{Tomaszewska2019fastcharge}, with a continuous taper:
\begin{itemize}
\item \textbf{CC phase} (SoC below the knee, $\text{SoC}_{\text{knee}}$=0.80): energy is delivered at constant
  power $P^{\text{eff}}_{ij}$, so SoC rises linearly.
\item \textbf{CV phase} (SoC $\geq \text{SoC}_{\text{knee}}$): power tapers linearly with the
  remaining headroom, $P(\text{SoC}) = P^{\text{eff}}_{ij}\,(1 - \text{SoC})/(1 -
  \text{SoC}_{\text{knee}})$, so SoC approaches full charge exponentially with time constant
  $\tau_{\text{CV}} = 0.20\, C_i / P^{\text{eff}}_{ij}$ ($P^{\text{eff}}_{ij}$ at the knee,
  ${\sim}0.5 P^{\text{eff}}_{ij}$ at 90\%, vanishing towards 100\%).
\end{itemize}
This continuous profile, rather than a two-step approximation, is what
makes the ``how much to charge'' decision non-trivial: the marginal minute
of charging buys progressively less energy above the knee, so the last
twenty points of charge can cost as much time as the previous sixty.

\subsection{Station congestion}
\label{sec:congestion}
We consider the general case in which stations may be \emph{shared}: vehicles
outside the fleet also use them, whether because charging takes place at public
stations or because a private operator runs other EVs on unrelated duties.  Either
way, fleet and non-fleet vehicles compete for the same connectors, so a vehicle
arriving to charge can find every EVSE busy and must wait.  A station's occupancy
(how many of its connectors are busy) is public: any vehicle can read it,
for instance from a standard status broadcast, and it is the only station
information a vehicle observes.  How heavy this external
demand is, and how it is modeled, are scenario choices (Section~\ref{sec:system}),
typically a first-in-first-out (FIFO) queueing discipline with mixed arrival and
service processes.

\subsection{Formal model}
\label{sec:formalmodel}
The problem is a decentralized partially observable decision process
(Dec-POMDP), the standard formalism for a team of agents that act on partial
local views of a shared state.  It is the tuple
\begin{equation*}
\langle\, \mathcal{I},\; \mathcal{S},\; \{\mathcal{A}_i\},\; P,\; R,\; \{\Omega_i\},\; O \,\rangle,
\end{equation*}
where:
\begin{itemize}
\item $\mathcal{I}$ is the set of $N$ vehicles, coupled through the station
  queues.
\item $\mathcal{S} = \big(\mathbf{c}, \{x_i\}_{i=1}^N\big)$ is the joint state: each vehicle's
  local state $x_i$ (its SoC, position and remaining route) together with the
  occupancy vector $\mathbf{c}$ over the $M$ stations.  No vehicle sees $\mathcal{S}$ in full:
  each observes only its own $x_i$ and the public occupancy $\mathbf{c}$
  (Section~\ref{sec:congestion}), not the private states of the others.
\item $\mathcal{A}_i$ is the action of vehicle $i$ at a decision point --- at each
  route stop, just before departing for the next: whether to charge, at which
  station, and to what target SoC.
\item $P$ is the state-transition rule, realized implicitly by the environment ---
  the simulator here, the real fleet in deployment: interleaved fleet actions,
  external arrivals, charging events and route progression.
\item $R$ is the team reward: a single episodic scalar for the whole fleet whose
  \emph{undiscounted} return the agent maximizes.  It operationalizes the objective
  (the success ratio) and its specific shaping is a design choice deferred to
  Section~\ref{sec:reward}.
\item $\Omega_i$ is vehicle $i$'s local observation space, and $O :
  \mathcal{S} \to \Omega_i$ the observation map that extracts vehicle $i$'s
  observation $o_i$ from the joint state: its own $x_i$ together with the public
  occupancies of the stations near it.
\end{itemize}
The environment, its dynamics $P$, the objective (maximizing the success ratio)
and the action set $\mathcal{A}_i$ (whether, where, how much) are the
\emph{problem}: fixed by the operational setting, whether real or simulated, they
hold for any controller.
Other choices (what signals vehicle $i$ reads through $O$, how the target SoC is
represented) are left open and deferred to the next section.

\section{Agent Design}
\label{sec:agent}

The agent is designed along three axes: (a) the signals each vehicle builds from
its local and public observations; (b) the rule that turns those signals into
charging decisions; and (c) the reward that scores the outcome.  The learning
algorithms and baselines that fill it in are the subject of
Section~\ref{sec:learning}.  Every vehicle runs the same policy (parameter sharing), acting on its own
observation $o_i$ (its private state plus the public occupancy signals of nearby stations).

At each decision point, vehicle~$i$ builds, for each station in its candidate set
$\mathcal{C}_i$ (the $K_{\text{ChS}}$ stations nearest its current position), a
block of four signals: two vehicle-side margins, shared across candidates, and two
station-side signals specific to each.  Together they combine the vehicle's own
state with the public occupancy broadcast and hold exactly what is needed to score
each candidate (Section~\ref{sec:action}).  These four signals are defined in
Section~\ref{sec:state}.

The action has two parts.  The first is which station to use: one candidate, or a
\emph{null} option under which the vehicle does not charge on the current leg (the
trip to the next stop).  The second, when it charges, is a target SoC, chosen in
steps of $\Delta_{\text{SoC}}$=0.05 (Table~\ref{tab:params}) from $0$ to $100\%$ and common to every solver:
\begin{equation}
\mathcal{A}_i = \left(\mathcal{C}_i \cup \{\varnothing\}\right) \times \Sigma,
\qquad \Sigma = \{0, 0.05, \dots, 1.00\}.
\label{eq:action_space}
\end{equation}

The charging decision decomposes naturally into two coupled subproblems (which
station to use and how much to charge) and we parametrize the policy to match.
The station-selection part assigns a preference score to each candidate,
reflecting that this stage is an evaluation of competing alternatives, while the
target-SoC part sets how much energy to add once a station is chosen.  This
decomposition embeds an inductive bias aligned with the problem structure, giving
a modular, interpretable policy and reducing the complexity each component must
learn.

A \emph{charging policy} $\pi_\theta \colon \Omega_i \to \mathcal{A}_i$, its
parameters $\theta$ shared across all vehicles, maps the observation $o_i$ to an
action by scoring each candidate and the \emph{null} option and picking the best.  We parametrize it in two
ways: \emph{per-candidate}, which scores each candidate on its own (Section~\ref{sec:action}), and
\emph{joint}, which scores all candidates together (Section~\ref{sec:pol_joint}).

\subsection{Decision Signals}
\label{sec:state}

At every decision point, vehicle~$i$ processes its observation $o_i$ (its own state together with the publicly broadcast occupancies of the nearby stations) into a four-signal block $\mathbf{s}_{ij}$ for each candidate station $j$:
\begin{equation}
\mathbf{s}_{ij} = \left[s_1,\; s_2,\; s_3,\; s_4\right]
\end{equation}
where the first two are vehicle-side margins (time and energy), shared by all candidates, and the last two combine the vehicle's travel times with publicly broadcast station information:
\begin{itemize}
\item \emph{Time margin}, $s_1 = (h_i + \delta_i - t) / H$: normalized time to the vehicle's deadline, with $h_i$ and $\delta_i$ as in Section~\ref{sec:problem} and $H$ the global time-budget cap used as a normalizer.
\item \emph{Energy margin}, $s_2$: the time the vehicle can keep driving before its SoC would fall to the floor $\text{SoC}_{\min}$, normalized by $H$.  It reaches $s_2$=1 when the current SoC is enough to finish the route.
\item \emph{Detour}, $s_3 = \max\!\big(0,\; (t_{i \to j} + t_{j \to \text{next}} - t_{i \to \text{next}}) / H\big)$: normalized extra travel time from diverting to station~$j$, at its publicly known location, instead of driving straight to the next stop.
\item \emph{Gain}, $s_4$: expected SoC gain if the vehicle stops at station~$j$ for a fixed reference time~$T_{\text{ref}}$, derived in Eqs.~\eqref{eq:kappa_ggk}--\eqref{eq:s4_def} below.
\end{itemize}

\paragraph{Derivation of $s_4$}
$s_4$ is the \emph{expected SoC gain in a reference window} $T_{\text{ref}}$: the charge the vehicle would add by stopping at station~$j$ for that window.  Two things shape it.  It follows the real, non-linear charging curve from the vehicle's current SoC, so a nearly full battery is credited with a smaller gain.  And it discounts the expected queueing time at the station, so a congested fast station can be worth less than a free slower one.  Both effects collapse into a single comparable number.

The signal is computed only from publicly available quantities: the station's nominal EVSE power
$P^{\text{evse}}_j$, its number of EVSE $E_j$, and a rolling estimate
$\hat{\rho}_j \in [0, 1)$ of its occupancy (the mean fraction of busy EVSE),
which each vehicle forms by sampling the station's public occupancy feed
(OCPP-style \texttt{StatusNotification} broadcasts). To
these the vehicle adds its own state: its BMS power cap $P^{\text{bms}}_i$,
current SoC $\text{SoC}_i$ and battery capacity $C_i$.

The sustained power the vehicle can draw is the effective power $P^{\text{eff}}_{ij}$ of
Eq.~\eqref{eq:power}, the station/vehicle minimum. When the station is busy, only part of the window is
spent charging and the rest is spent waiting.  We estimate the charging
fraction as the service-to-sojourn ratio $\kappa = t_{\text{serv}}/(t_{\text{serv}} + W_q)$,
taking $W_q$ from the Sakasegawa $G/G/k$ waiting-time
approximation~\citep{Sakasegawa1977approximation}, which, unlike an $M/M/c$ model,
assumes no particular arrival or service distribution and needs only the mean
occupancy $\hat{\rho}_j$ and the number of EVSE $E_j$, not the full arrival or
service-time distributions.  This is deliberate: $E_j$ is known static station data and
$\hat{\rho}_j$ is exactly what the operator broadcasts.  The resulting fraction $\kappa$, the time spent charging over the total time at the station (charging plus queueing), is

\begin{equation}
\kappa(\hat{\rho}_j, E_j) \;=\; \frac{E_j (1 - \hat{\rho}_j)}{E_j (1 - \hat{\rho}_j) + \hat{\rho}_j^{\,E_j}},
\label{eq:kappa_ggk}
\end{equation}

which tends to $1$ for an empty station and to $0$ as $\hat{\rho}_j \to 1$,
with a sharper drop as $E_j$ grows.
We use the integer exponent $E_j$ in place of Sakasegawa's exact exponent
$\sqrt{2(E_j+1)}-1$ as a lightweight simplification: the two coincide at $E_j$=1,
where $\kappa = 1 - \hat{\rho}_j$ is exact, and diverge only mildly for larger $E_j$;
as a decision feature, $\kappa$ need not be exact.  The mean charging time within the window
is then $t_{\text{eff}} = \kappa(\hat{\rho}_j, E_j)\, T_{\text{ref}}$.

Projecting this time onto the CC/CV curve of Section~\ref{sec:cccv}, from the current SoC $\text{SoC}_i$ at power $P^{\text{eff}}_{ij}$, gives the SoC the vehicle would reach:

\begin{equation}
\text{SoC}_i(t_{\text{eff}}) \;=\;
\begin{cases}
\text{SoC}_i + \frac{P^{\text{eff}}_{ij}\, t_{\text{eff}}}{C_i}
   & \text{if CC ends within the window,}\\[4pt]
1 - (1 - \text{SoC}_\text{knee})\, \exp\!\left(-(t_{\text{eff}} - t_\text{knee})/\tau_{\text{CV}}\right)
   & \text{if the window crosses the knee,}\\[4pt]
1 - (1 - \text{SoC}_i)\, \exp\!\left(-t_{\text{eff}}/\tau_{\text{CV}}\right)
   & \text{if } \text{SoC}_i \geq \text{SoC}_\text{knee},
\end{cases}
\label{eq:soc_proj}
\end{equation}

where $t_\text{knee} = (\text{SoC}_\text{knee} - \text{SoC}_i)\, C_i / P^{\text{eff}}_{ij}$ is the time to reach the knee.  The signal is the resulting gain, capped at the battery's usable maximum $\text{SoC}_{\max}$=0.99:
\begin{equation}
s_4 \;=\; \mathrm{clip}\!\left( \min(\text{SoC}_i(t_{\text{eff}}), \text{SoC}_{\max}) - \text{SoC}_i, \; 0,\, 1 \right).
\label{eq:s4_def}
\end{equation}

We set $T_{\text{ref}}$=0.5~h, a typical mid-shift charging dwell.  It is long enough to separate station types (a 22~kW AC unit, a 50~kW DC unit, a 150~kW fast charger) and short enough to stay informative when the controller decides per stop.

Let us remark that $s_4$ is a reference, not an exact figure.  The real charging time is not known in advance, so a fixed window stands in for it, and the gain is computed from the vehicle's current SoC, not the lower SoC it will have after the detour to the station.  Neither approximation has to be exact.  The detour is already in $s_3$, the time and energy margins in $s_1, s_2$, and the policy learns to combine the four signals and absorb the slack.

Rather than exposing every low-level variable on its own, we encode them into decision-oriented signals that capture the factors governing station selection.  This builds domain knowledge into the state representation, letting the policy focus on learning the station-selection strategy rather than rediscovering well-established queueing and charging dynamics.

\subsection{Per-candidate policy}
\label{sec:action}

The per-candidate policy $\pi_\theta$ is made of two functions of a candidate's four-signal block, produced as the two outputs of a single shared network (hence the common parameters $\theta$): a score $u_\theta$ that ranks candidates, and a target-SoC map $\sigma_\theta$ that says how much to charge.  At each decision point, vehicle~$i$ scores each of its $|\mathcal{C}_i|$ candidates with $u_\theta$, together with a synthetic \emph{null} option $\mathbf{s}_{i\varnothing} = [s_1, s_2, 0, 0]$, the choice not to charge, with no detour ($s_3$=0) and no gain ($s_4$=0), and charges at the top-scoring station, or not at all if the \emph{null} wins:
\begin{equation}
j^* = \argmax_{j \in \mathcal{C}_i \cup \{\varnothing\}} \; u_\theta(\mathbf{s}_{ij}),
\qquad
a_i = \begin{cases}
\text{no charge} & j^* = \varnothing\\
\text{charge at } j^* & j^* \in \mathcal{C}_i
\end{cases}
\label{eq:station_select}
\end{equation}
where $a_i$ is the action.  When a real station wins, $\sigma_\theta(\mathbf{s}_{ij^*}) \in \Sigma$ sets its target SoC (Eq.~\ref{eq:action_space}); the \emph{null} carries none.

Because the \emph{null} competes as one more candidate, ``charge or not'' needs no hand-tuned threshold.  The policy learns when charging beats waiting, and the answer shifts with context: the urgency in $s_1, s_2$ and the detour and congestion of the candidates in $s_3, s_4$.

\subsection{Joint policy}
\label{sec:pol_joint}
The joint policy applies the same idea, but scoring all candidates at once instead of one at a time.  A single network $g_\theta$ reads the shared vehicle signals $s_1, s_2$ and the station signals $(s_3, s_4)$ of every candidate, and scores them all in one pass:
\begin{equation}
\big(u_{i\varnothing},\, u_{i1},\, \dots,\, u_{i|\mathcal{C}_i|}\big) \;=\; g_\theta\big(s_1,\, s_2,\, (s_3,s_4)_1,\, \dots,\, (s_3,s_4)_{|\mathcal{C}_i|}\big),
\label{eq:joint_policy}
\end{equation}
Unlike the per-candidate form, the \emph{null} is not a synthetic input block here but a dedicated output, the score $u_{i\varnothing}$.  The station then follows from the same argmax-with-null of Eq.~\eqref{eq:station_select}, and the target SoC is one further output of the same network $g_\theta$: the joint form is thus a single network with $|\mathcal{C}_i|+1$ station scores and one target-SoC output.

The contrast with the per-candidate form is deliberate.  The per-candidate
policy shares one scoring function across candidates, a structural prior of
parameter sharing and permutation invariance.  The joint form drops that prior
and lets the optimizer read all candidates at once.  Comparing the two isolates
how much of the coordination comes from the prior and how much the optimizer
discovers on its own (Section~\ref{sec:experiments}).

\subsection{Reward}
\label{sec:reward}
The success ratio of Section~\ref{sec:problem} can be engineered as a team reward, one scalar
per episode.  Within an episode, a vehicle scores $+1$ for completing its route.  A failure, battery
below $\text{SoC}_{\min}$ or a timeout past its time budget $h_i$, scores $-1$, raised by
a small per-leg credit $c \ll 1$ for the progress made before failing:
\begin{equation}
r_i = \begin{cases} +1 & \text{route completed,}\\[2pt] -1 + c\,\ell_i & \text{failed after } \ell_i \text{ legs,}\end{cases}
\qquad
R = \frac{1}{N}\sum_{i=1}^{N} r_i \;\in\; [-1, 1].
\label{eq:reward}
\end{equation}
The episode reward $R$ is the fleet average, reading as the fraction of the fleet
that succeeds minus the fraction that fails.  Training and evaluation average $R$
over $J$ independent episodes.

\section{Learning Algorithms and Reference Controllers}
\label{sec:learning}
We search the charging policy with two learners, NEAT and PPO
(Sections~\ref{sec:neat} and~\ref{sec:ppo}).  Beside them we study a
congestion-blind simulated-annealing (SA) control (Section~\ref{sec:sa_heuristic}), and measure them all
against reference controllers: an omniscient Oracle, a \emph{greedy} rule and a \emph{null}
policy (Sections~\ref{sec:oracle} and~\ref{sec:baselines}).  Every controller here except the
Oracle is \emph{reactive}: it decides at each stop from the signals available then, whereas the Oracle plans the whole shift in advance.  As
Section~\ref{sec:res_base} will show, NEAT, PPO and even the congestion-blind SA far outperform
the baselines, and PPO comes close to the omniscient Oracle.

\subsection{NEAT (Neuroevolution)}
\label{sec:neat}
NEAT~\citep{Stanley2002neat} learns the policy by evolving a population of neural networks.  Each network starts simple and grows over generations: mutations add nodes and connections and perturb the weights, the fitter networks reproduce, and \emph{speciation} shelters new structures long enough to prove themselves.  Within our paradigm, the network maps the signals to candidate scores and the target SoC, in either of the two policy forms of Section~\ref{sec:agent}: a per-candidate scorer of fixed size ($4\to2$), applied in turn to each of the $|\mathcal{C}_i|$ candidates and the \emph{null}, or a joint network whose size grows with the candidate count $|\mathcal{C}_i|$, the two NEAT rows of Table~\ref{tab:instantiations}.

Being gradient-free, NEAT scores each whole episode as one fitness value and never attributes credit to individual decisions, which suits the noisy, episodic reward of fleet simulation.  Evolution is by now an established alternative to gradient RL at this scale~\citep{Salimans2017openai, Such2017uber, TensorNEAT2025}, yet almost unused for EV charging: its one prior use tunes charging current at a single workplace site, without station selection~\citep{Kemper2025neat}, far from the fleet-scale problem here~\citep{Wu2024rlsurvey}.  The reference settings used in the evaluation of Section~\ref{sec:experiments} are given in Table~\ref{tab:hparams}.

\subsection{PPO (Proximal Policy Optimization)}
\label{sec:ppo}
PPO~\citep{Schulman2017ppo} is a standard policy-gradient method from reinforcement
learning: it improves the policy in small steps that stay close to the current one
(its clipped surrogate objective), guided by generalized advantage estimation (GAE) from a learned value
function.  We train it on the same observation and simulator
as NEAT, acting in the same space (Eq.~\ref{eq:action_space}), where the station
choice and the target SoC are each drawn from a categorical distribution.  Its
hyperparameters are collected in Table~\ref{tab:hparams}, including a discount
$\gamma$=0.99.  Although the objective is undiscounted (Section~\ref{sec:agent}),
PPO computes its updates from a discounted return, and over single-shift episodes this
makes no practical difference.

The fleet is a multi-agent system, but PPO learns from single-agent trajectories in which each step is the consequence of the previous action.  The fleet breaks this in two ways.  First, the shifts it runs are independent episodes (Section~\ref{sec:problem}), so a training trajectory must not span two of them.  Each PPO training episode is therefore limited to one shift.  Second, all $N$ vehicles share one policy (parameter sharing)~\citep{Terry2020paramsharing}, and within a shift their decisions interleave in time, so consecutive steps typically belong to different, unrelated vehicles.

Left alone, this is the \emph{independent-learning} setup, a shared-parameter PPO per vehicle, which leaves the interleaved credit assignment ill-posed under an episodic reward~\citep{Foerster2018coma, Dulac2021challenges} and does not train.  The standard remedy is \emph{centralized training with decentralized execution}, a joint-state critic (MAPPO)~\citep{Yu2022mappo}, but its policy-gradient variance grows with the number of agents~\citep{Kuba2021variance}.  We tried both across per-vehicle reward attributions and rollout schemes, and over a hyperparameter sweep.  However, none reached the level of the single-agent approach we describe next, consistent with these known difficulties~\citep{Majid2024drl_vs_es, Li2025tevc_survey}.

Our approach reduces training to a single \emph{ego-learner}.  Each episode follows one learner vehicle, whose decisions form a clean single-agent trajectory, while the other $N-1$ run a \emph{frozen} policy held fixed through the episode, so the learner faces a stationary environment.  That frozen policy is not always the most recent one: it is drawn from a pool of past checkpoints (refreshed every $K_{\text{frozen}}$ rounds, Table~\ref{tab:hparams}), so the learner trains against a range of past fleet behaviors rather than a single one.
Sampling from a pool keeps the learner from overfitting to exploit one frozen opponent and gives it a steadier, more varied set of rivals.  Training a policy against past versions of itself is a form of \emph{self-play}~\citep{Heinrich2016fsp}, and here it stabilizes learning.  The training signal is the learner's own share of the team reward (Eq.~\ref{eq:reward}).  This structure decouples the learner's policy from the rest of the fleet only during training.  Training yields a single policy, which every vehicle then runs at deployment.

One further mechanism proved essential across repeated training runs: a \emph{behavioral-cloning warm-start}.  We clone the policy of the simple \emph{greedy} rule (Section~\ref{sec:baselines}) into the actor (${\sim}20$k state--action pairs, fitted by cross-entropy), so PPO starts from that rule's level rather than from scratch.  Without it, random exploration charges needlessly and drops below the \emph{null} controller (Section~\ref{sec:baselines}) --- worse than never charging at all --- before the episodic gradient recovers.

PPO runs in the same two policy forms as NEAT, the per-candidate scorer of Eq.~\eqref{eq:station_select} and the joint network of Eq.~\eqref{eq:joint_policy}, here as networks of fixed architecture (the two PPO rows of Table~\ref{tab:instantiations}).

Taken together, PPO needs a much heavier training setup than NEAT: the single-agent reduction, the behavioral-cloning warm-start and the historical pool are all needed before it trains at all, where NEAT runs the same policy search without any of them.  This gap is itself a result, and one of the reasons we tried NEAT as one of the control algorithms in the first place.  Nonetheless, PPO is a useful state-of-the-art reference~\citep{Schulman2017ppo} and has been applied at fleet scale~\citep{Alqahtani2022marl}, so we evaluate it in full.  Despite the heavier setup, it proves the strongest controller of all (Section~\ref{sec:res_base}).

\begin{table}[t]
\centering
\caption{The four optimizer--structure controllers.  Sizes are inputs\,$\to$\,outputs; the per-candidate size is fixed while the joint size grows with the candidate count $|\mathcal{C}_i|$ (realized by a network fixed to the cap $K_{\text{ChS}}$=5, unused slots zero-padded).  Every target-SoC output ranges over the shared grid $\Sigma$ (Eq.~\ref{eq:action_space}), and PPO network widths are in Table~\ref{tab:hparams}.}
\label{tab:instantiations}
\small
\begin{tabular}{@{}lll@{}}
\toprule
\textbf{Controller} & \textbf{Reads candidates} & \textbf{Network} \\
\midrule
NEAT-per-candidate & each of $|\mathcal{C}_i|$ blocks $+$ \emph{null} & evolved topology, $4\to2$ per block (score, SoC) \\
NEAT-joint         & all at once, \emph{null} an output     & evolved topology, $(2{+}2|\mathcal{C}_i|)\to(|\mathcal{C}_i|{+}1)$ logits, SoC \\
PPO-per-candidate  & each of $|\mathcal{C}_i|$ blocks $+$ \emph{null}  & fixed net, $4\to2$ per block (score, SoC) \\
PPO-joint          & all at once, \emph{null} an output     & fixed net, $(2{+}2|\mathcal{C}_i|)\to(|\mathcal{C}_i|{+}1)$ logits, SoC \\
\bottomrule
\end{tabular}
\end{table}

\subsection{A congestion-blind control: the SA-tuned rule}
\label{sec:sa_heuristic}
The SA-tuned rule is a threshold controller of the same family as the two
learners, tuned on the same simulator and reward, but reading only a
reduced set of signals: the vehicle's own urgency, its state of charge and its
projected charge at route completion, with no occupancy and no detour.  The rule
is straightforward: it charges only when the battery looks short on both counts,
its current SoC below a threshold \emph{and} its SoC projected at route
completion below the reserve,
\begin{equation}
\text{SoC}_i \;<\; \theta_{\text{SoC}}
\qquad\text{and}\qquad
\text{SoC}_i - \zeta_i\, d_{\text{remaining}} \;<\; \text{SoC}_{\min},
\label{eq:sa_trigger}
\end{equation}
with $d_{\text{remaining}}$ the route distance still to cover; it then drives to
the nearest station and charges to
\begin{equation}
\text{SoC}_{\text{target}} = \min\left(\theta_{\text{max}},\; \zeta_i \cdot d_{\text{remaining}} + \text{SoC}_{\min} + \theta_{\text{margin}}\right).
\end{equation}
Its three parameters $(\theta_{\text{SoC}}, \theta_{\text{max}},
\theta_{\text{margin}})$, with search ranges in Table~\ref{tab:hparams}, are
tuned by simulated annealing on the same fleet reward (Eq.~\ref{eq:reward}).

This control serves two purposes: it shows how far the paradigm alone carries a
controller, and, by dropping only the occupancy and detour signals, it bounds
what congestion-awareness is worth, the gap to the learners being an upper bound
on the combined value of those signals.  Threshold rules like this are in any
case a common way to run electric fleets~\citep{Chen2016saev, Loeb2019saev},
alongside priority rules for scheduling~\citep{Xu2016priority}, so even a tuned
instance is a controller in its own right, and it does well wherever congestion
stays light, the gap to the learned controllers opening only where congestion
bites (Section~\ref{sec:res_base}).

\subsection{The Oracle}
\label{sec:oracle}
The \emph{Oracle} sets the performance ceiling for the learners: an omniscient planner that knows an episode in full (every route, the fleet parameters and the realized external congestion) and searches offline for the best charging plan for that exact shift.  A plan assigns each vehicle its charges, each defined by the route leg at which the vehicle detours to charge, the station, and the target SoC.  To score a plan, the Oracle runs the real simulator (Section~\ref{sec:system}) with it and takes the resulting fleet reward, exactly as the learners do.

The Oracle is warm-started from the best reactive controller on each city: that controller's realized charging decisions form the initial plan, which simulated annealing then polishes.  At each step the Oracle perturbs the current plan --- most often moving a charge to another station or leg, or nudging a target SoC, and less often adding, removing, swapping across vehicles or reordering charges (Table~\ref{tab:oracle}) --- and accepts it by the Metropolis rule: always when the new plan scores better, and otherwise with a probability that shrinks as the temperature cools on a geometric schedule.  Route feasibility (finishing on time and above the reserve) is enforced by the reward itself, with no slack variables.  The schedule is intentionally deep: a slow geometric cooling ($\alpha$=0.995) down to $T_{\min}$=$10^{-9}$, with $90$ plan evaluations per temperature and $20\,000$ evaluations in all, kept as the best of five independent restarts (Table~\ref{tab:oracle}).

The Oracle needs the future, so it is not a deployable policy, and its offline search is computationally heavy.  Even so, the Oracle serves as a strong planning reference that makes each controller's remaining gap interpretable.  The benchmark scenarios are chosen so the Oracle resolves each in the $99$--$100\%$ band (Section~\ref{sec:benchmark}): demanding but attainable, with no slack to spare.  Using a full-information planner
as a benchmark is standard for charging controllers: the EV2Gym benchmark, for one,
includes a mixed-integer optimization oracle for the same role~\citep{Orfanoudakis2025ev2gym}.

\begin{table}[t]
\centering
\caption{Oracle meta-optimizer: simulated annealing over per-vehicle charging plans.  A plan gives each vehicle at most $s$ charges (capped-$s$; $s$=8 here).  Each step applies one perturbation and is scored on the real simulator; the search is warm-started from the best reactive controller on each city and reports the best of five independent restarts.}
\label{tab:oracle}
\small
\begin{tabular}{@{}lrl@{}}
\toprule
\textbf{Perturbation} & \textbf{Prob.} & \textbf{Schedule and limits} \\
\midrule
move a charge to another station     & $0.30$ & restarts $N_{\text{restart}}$: $5$ (best-of) \\
move a charge to another leg         & $0.20$ & evaluations per temperature: $90$ \\
nudge target SoC ($\pm0.05,\pm0.10$)  & $0.20$ & total plan evaluations: $20\,000$ \\
add a charge                         & $0.15$ & cooling: geometric, $\alpha$=0.995 \\
remove a charge                      & $0.10$ & $T_{\min}$=$10^{-9}$ \\
swap a charge across vehicles        & $0.04$ & charge cap $s$: $8$ \\
reorder charges within a vehicle     & $0.01$ & warm start: best reactive controller \\
\bottomrule
\end{tabular}
\end{table}

\subsection{Rule-based baselines}
\label{sec:baselines}
Two rule-based controllers set the lower end:
\begin{itemize}
\item \emph{greedy}: whenever its SoC drops below $20\%$, the vehicle drives to the nearest station and charges to $100\%$, using nothing but its own position and SoC.
\item \emph{null}: never charges.
\end{itemize}

\emph{null} is ineffective, but not the worst a policy can do.  Charging badly, for instance charging at every stop, wastes so much time that vehicles never finish their routes and always time out.  By the reward of Eq.~\eqref{eq:reward}, a fleet in which no vehicle completes scores the minimum, $R$=$-1$: that, not \emph{null}, is the true lower bound on any policy.


\section{The Evaluation Environment}
\label{sec:system}

\begin{figure*}[!tbp]
\centering
\includegraphics[width=0.62\textwidth]{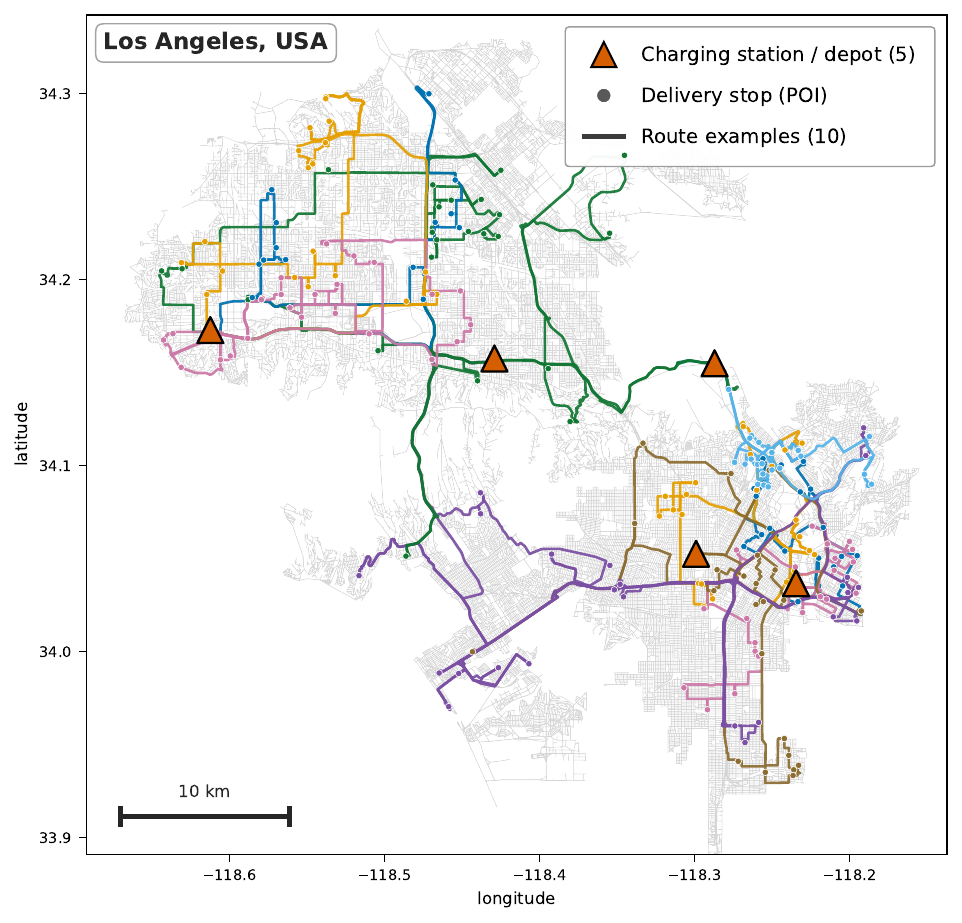}
\caption{An illustrative deployment scenario (Los Angeles).  The city's real
OpenStreetMap road network underlies one sampled scenario: five
shared charging stations (triangles), drawn from the graph, and a fleet of $45$ delivery vehicles whose
stops are sampled from ${\sim}5000$ candidate points of interest.  Each station
doubles as a depot, so a route starts and ends at the triangle that serves it.
Ten example routes are shown (colored lines), with dots marking the shift's stops
and the scale bar the county-scale extent.  This is the static layout the
controllers act on.  Charging decisions, battery state and timing appear only
once the scenario is executed (Fig.~\ref{fig:trace_subdivide}(a) and~(b)).}
\label{fig:scenario_example}
\end{figure*}

The controllers are trained and evaluated in a discrete-event simulator that
realizes the operational problem of Section~\ref{sec:problem} on real urban road
networks.  Each city's drivable graph is imported from OpenStreetMap via
OSMnx~\citep{Boeing2017osmnx}, which supplies edge lengths, speed limits and
travel times.  Charging stations, depots and delivery points all sit on nodes of
this graph, and the delivery points are sampled uniformly over the nodes, so they
concentrate where the road network is denser, a proxy for the built-up parts
of the city, rather than spreading evenly over the map, which keeps the routes
realistic.  Figure~\ref{fig:scenario_example} shows one frozen scenario.

\subsection{Scenario, Fleet and External Demand}

At the start of every simulated shift the environment draws a fresh fleet and
scenario (per-vehicle parameters, station positions and configuration, and
external demand), each sampled independently from the distributions of
Table~\ref{tab:params}, so no two shifts are identical.
By randomizing the scenario each shift, we expose the policy to diverse city configurations, so it learns transferable decision rules, rather than specializing to one city (domain randomization, Section~\ref{sec:experiments}).  In the evaluation benchmark, one such draw is frozen per
city (Section~\ref{sec:benchmark}).

We model external demand at each station as a Poisson process of rate
$\lambda_{\text{ext}}$, with each external user occupying a connector for an exponentially
distributed time of mean $\mu_{\text{ext}}^{-1}$.  This is a finite-capacity
M/M/$E_j$/$K$ queue: external arrivals that find the station full leave without charging.  Such models are standard for charging-station
demand~\citep{Bae2012demand}.

As introduced in Section~\ref{sec:problem}, fleet and external vehicles share the connectors.
Those that find every connector
busy join a single FIFO queue, served in arrival order, so external
users take connectors the fleet would otherwise use and increase its waiting time.
This reproduces the temporal correlation of real congestion, unlike an i.i.d.\ model
in which each connector is independently busy with fixed probability. Fleet
vehicles charge under the CC/CV physics of Section~\ref{sec:physics}, with power
shared among simultaneous users, so the aggregate (external plus fleet) is not a homogeneous
exponential-service queue.  The simulator tracks each station's occupancy as the fraction of its $E_j$ connectors that are busy, and the rolling mean of this fraction, $\hat{\rho}_j$, is the congestion input to the gain signal $s_4$ (Section~\ref{sec:state}).

\subsection{Route Generation}

Routes follow a time-budget model, with all its ranges in
Table~\ref{tab:params}.  For each vehicle and working shift we sample a time budget
$h_i$.  Starting from the vehicle's home station, delivery points are
added one by one by a radius rule: with probability $p_{\text{jump}}$ the next
point is drawn uniformly within a long-range radius $r_L$ (inter-neighborhood
trips), otherwise within a fixed short-range radius $r_S$ (local deliveries).
When fewer than three candidates fall inside the radius, a nearest-neighbor
fallback is used.  Stops are added until the estimated total time (travel,
service, the return leg to the depot, and the reserve $\varphi\,h_i$ of
Section~\ref{sec:problem}) exceeds $h_i$.  Service times (the time a vehicle spends stopped at each delivery) are
exponential with mean $\bar{t}_{\text{svc}}$.  The time budget, the jump probability, the
long radius and the service-time mean are each resampled per route, and only the
short radius is fixed.

\subsection{Discrete-Event Simulation Engine}

The simulator advances a concrete realization of the abstract state $\mathcal{S}$ of Section~\ref{sec:formalmodel}.
For each vehicle, it tracks the state of charge, the position along the route, the stops still to serve and the time left in the shift.  For each station, it tracks which connectors are busy, its FIFO queue, and every active charging session with its target SoC and its CC/CV phase.  Three event types advance this state in temporal order:
\begin{itemize}
\item \textbf{LEG}: A vehicle completes a route segment and is about to start the next. The controller is queried to decide whether to continue the route or divert to a charging station.
\item \textbf{CHARGE}: A vehicle arrives at a charging station. It is either assigned a connector or placed in the queue.
\item \textbf{SLOT}: A periodic event (every $\Delta t_{\text{slot}}$) that updates all active charging sessions, promotes queued vehicles, and refreshes external occupancy.
\end{itemize}

A vehicle's shift ends in one of the three ways of Section~\ref{sec:problem} (on-time completion, battery depletion, or shift timeout) and its per-vehicle reward $r_i$ enters the episode reward $R$ of Eq.~\eqref{eq:reward}.

\subsection{Statistical Routing Model}
\label{sec:stat_model}

Computing an exact route on the city graph is expensive: for instance, a single shortest-path
query on the Los Angeles network (${\sim}49$k nodes) takes about $90$~ms.  A
training run issues one or more such queries for every charging decision,
vehicle, episode and generation, so exact routing is infeasible at this scale.
We therefore replace it with a per-city, stationary statistical surrogate that
answers each query in about $1$~\textmu s, five orders of magnitude faster,
at a small, measured loss of accuracy.

The surrogate is calibrated once per city.  For each training and benchmark city
we download its drivable network via OSMnx~\citep{Boeing2017osmnx} and sample
$10\,000$ random origin--destination pairs.  For each pair we record the
Euclidean (haversine) distance $d_{\text{euc}}$, the shortest-path road distance
$d_{\text{road}}$ and the travel time $t$, and derive the
\emph{circuity}~\citep{Boeing2017osmnx} $\tau = d_{\text{road}}/d_{\text{euc}}$
and the \emph{effective speed} $v = d_{\text{road}}/t$.  Pairs are bucketed by
$d_{\text{euc}}$ into 29 log-spaced bins over $[0.1, 100]$~km, and the per-bin
means $\bar{\tau}_b, \bar{v}_b$ are stored as that city's calibration.  At query
time the surrogate returns $d_{\text{road}} = d_{\text{euc}} \cdot \bar{\tau}_b$
and $t = d_{\text{road}} / \bar{v}_b$ deterministically, so the controller's
on-board predictions and the simulator's executions are bit-identical functions
of the endpoints and the city.  The bin edges are common to all cities, and only
the per-bin $(\bar{\tau}_b, \bar{v}_b)$ are city-specific, so each city keeps its
own distance-dependent circuity and speed.  Because the per-bin means do not form
a metric, the surrogate does not guarantee the triangle inequality across bins,
so the detour signal $s_3$ clips negative values to zero (Section~\ref{sec:state}).

\begin{figure*}[t]
\centering
\begin{tikzpicture}
\definecolor{cMad}{HTML}{0072B2}      
\definecolor{cLA}{HTML}{D55E00}       
\definecolor{cSyd}{HTML}{009E73}      
\definecolor{cTor}{HTML}{CC79A7}      

\begin{groupplot}[
    group style={group size=2 by 2, horizontal sep=1.6cm, vertical sep=1.7cm},
    width=0.46\textwidth,
    height=4.6cm,
    grid=major,
    grid style={gray!15},
    tick label style={font=\footnotesize},
    label style={font=\small},
    title style={font=\small\bfseries, yshift=-2pt},
    xmode=log,
    xmin=0.1, xmax=100,
    log basis x={10},
    every axis plot/.append style={line width=1.2pt},
    no marks,
]

\nextgroupplot[ylabel={Circuity $\tau$}, title={(a) Circuity vs $d_{\text{euc}}$},
               ymin=1.0, ymax=3.0,
               legend pos=north east,
               legend style={font=\scriptsize, draw=gray!40, inner sep=2pt}]
\addplot[color=cMad, solid]      table[col sep=comma, x=d_euc_center, y=circuity_mean]
                                 {data_routing_madrid_cal.csv};
\addlegendentry{Madrid}
\addplot[color=cLA, dashed]      table[col sep=comma, x=d_euc_center, y=circuity_mean]
                                 {data_routing_la_cal.csv};
\addlegendentry{Los Angeles}
\addplot[color=cSyd, densely dashdotted] table[col sep=comma, x=d_euc_center, y=circuity_mean]
                                 {data_routing_sydney_cal.csv};
\addlegendentry{Sydney}
\addplot[color=cTor, densely dotted] table[col sep=comma, x=d_euc_center, y=circuity_mean]
                                 {data_routing_toronto_cal.csv};
\addlegendentry{Toronto}

\nextgroupplot[ylabel={Effective speed (km/h)}, title={(b) Speed vs $d_{\text{euc}}$},
               ymin=0, ymax=100]
\addplot[color=cMad, solid]      table[col sep=comma, x=d_euc_center, y=speed_mean]
                                 {data_routing_madrid_cal.csv};
\addplot[color=cLA, dashed]      table[col sep=comma, x=d_euc_center, y=speed_mean]
                                 {data_routing_la_cal.csv};
\addplot[color=cSyd, densely dashdotted] table[col sep=comma, x=d_euc_center, y=speed_mean]
                                 {data_routing_sydney_cal.csv};
\addplot[color=cTor, densely dotted] table[col sep=comma, x=d_euc_center, y=speed_mean]
                                 {data_routing_toronto_cal.csv};

\nextgroupplot[xlabel={Actual $d_{\text{road}}$ (km)},
               ylabel={Predicted $d_{\text{road}}$ (km)},
               title={(c) Distance fit (held-out 2k)},
               xmode=normal, ymode=normal,
               xmin=0, xmax=120, ymin=0, ymax=120,
               every axis plot/.append style={line width=0.0pt, only marks, mark size=0.5pt}]
\addplot[color=cMad, mark=*, mark options={fill=cMad}]
    table[col sep=comma, x=d_road_actual, y=d_road_pred] {data_routing_madrid_fit.csv};
\addplot[color=cLA, mark=square*, mark options={fill=cLA}]
    table[col sep=comma, x=d_road_actual, y=d_road_pred] {data_routing_la_fit.csv};
\addplot[color=cSyd, mark=triangle*, mark options={fill=cSyd}]
    table[col sep=comma, x=d_road_actual, y=d_road_pred] {data_routing_sydney_fit.csv};
\addplot[color=cTor, mark=diamond*, mark options={fill=cTor}]
    table[col sep=comma, x=d_road_actual, y=d_road_pred] {data_routing_toronto_fit.csv};
\addplot[color=black, dashed, line width=1.1pt, no marks, forget plot] coordinates {(0,0) (120,120)};

\nextgroupplot[xlabel={Actual travel time (h)},
               ylabel={Predicted travel time (h)},
               title={(d) Travel time fit (held-out 2k)},
               xmode=normal, ymode=normal,
               xmin=0, xmax=3, ymin=0, ymax=3,
               every axis plot/.append style={line width=0.0pt, only marks, mark size=0.5pt}]
\addplot[color=cMad, mark=*, mark options={fill=cMad}]
    table[col sep=comma, x=t_travel_actual, y=t_travel_pred] {data_routing_madrid_fit.csv};
\addplot[color=cLA, mark=square*, mark options={fill=cLA}]
    table[col sep=comma, x=t_travel_actual, y=t_travel_pred] {data_routing_la_fit.csv};
\addplot[color=cSyd, mark=triangle*, mark options={fill=cSyd}]
    table[col sep=comma, x=t_travel_actual, y=t_travel_pred] {data_routing_sydney_fit.csv};
\addplot[color=cTor, mark=diamond*, mark options={fill=cTor}]
    table[col sep=comma, x=t_travel_actual, y=t_travel_pred] {data_routing_toronto_fit.csv};
\addplot[color=black, dashed, line width=1.1pt, no marks, forget plot] coordinates {(0,0) (3,3)};

\end{groupplot}
\end{tikzpicture}
\caption{Per-city statistical routing model calibrated on four cities (Section~\ref{sec:benchmark}): two used for training (Madrid, Los Angeles) and two held out for the benchmark (Sydney, Toronto).  (a)~Per-bin mean circuity $\tau = d_{\text{road}}/d_{\text{euc}}$ and (b)~effective speed $v = d_{\text{road}}/t$, each computed from 10\,000 OD pairs sampled from the city's OSMnx drivable network and binned into 29 log-spaced bins of $d_{\text{euc}} \in [0.1, 100]$~km.  (c,~d)~Predicted (bin-mean) vs actual distance and travel time on a held-out validation set of 2\,000 OD pairs per city; the diagonal marks a perfect fit, and the corresponding $R^2$ values are reported in Table~\ref{tab:routing_R2}.}
\label{fig:routing_per_city}
\end{figure*}
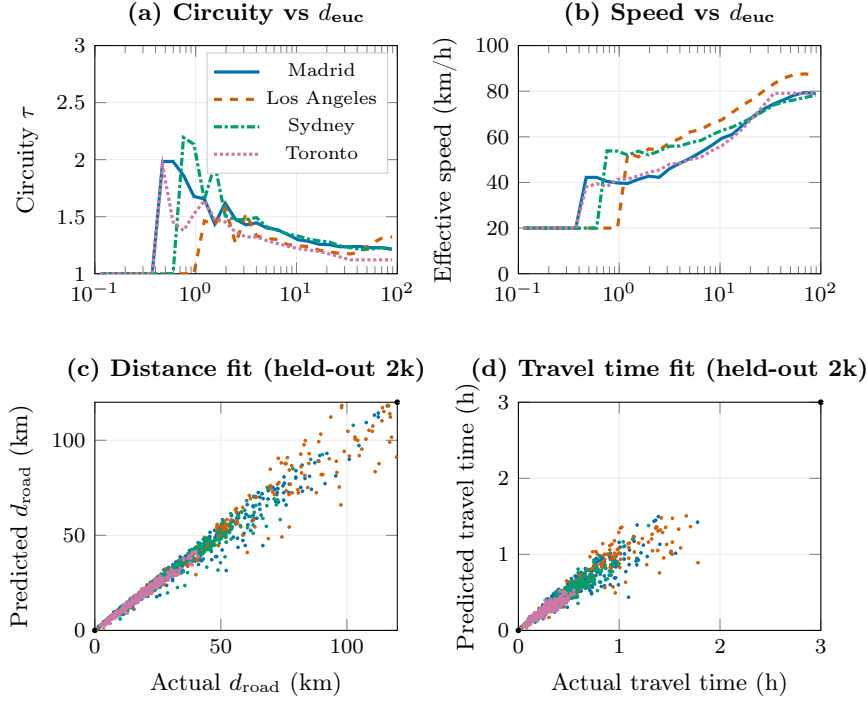

Per short leg the surrogate's relative error is around $20\%$ in distance and
$30\%$ in time, but these errors cancel rather than accumulate: over a full route
(${\sim}33$ legs) it estimates total distance to a mean error of $7.6\%$ (median
$4.6\%$) and total time to $9.3\%$ (median $7.1\%$), with only a small downward
bias in distance (${\sim}4\%$) and essentially none in time.  On held-out
origin--destination pairs, its calibration reproduces $93$--$98\%$ of the
variance in road distance and $85$--$92\%$ in travel time, the coefficient of
determination $R^2$, the fraction of the real variation the model captures.
Figure~\ref{fig:routing_per_city} shows the per-city fits and
Table~\ref{tab:routing_R2} the per-city $R^2$ and mean absolute error.  Crucially, every controller and the
Oracle run on the same surrogate under common random numbers, so this residual
error is shared identically and does not affect the comparison: it changes only
the absolute realism, not the ranking.

\begin{table}[tb]
\centering
\caption{Validation of the per-city routing surrogate on $2\,000$ held-out OD pairs per city.  $R^2$ and mean absolute error (MAE) for road distance ($d_{\text{road}}$, km) and travel time ($t$, h).}
\label{tab:routing_R2}
\small
\begin{tabular}{lrrrr}
\toprule
City & $R^2$($d_{\text{road}}$) & $R^2$($t$) & MAE($d_{\text{road}}$) [km] & MAE($t$) [h] \\
\midrule
Madrid (Spain)        & 0.972 & 0.864 & 2.66 & 0.075 \\
Los Angeles (USA)     & 0.929 & 0.847 & 4.78 & 0.083 \\
Sydney (Australia)    & 0.964 & 0.921 & 2.16 & 0.045 \\
Toronto (Canada)      & 0.980 & 0.856 & 0.95 & 0.033 \\
\bottomrule
\end{tabular}
\end{table}

\section{Experiments and Results}
\label{sec:experiments}

The controllers of Section~\ref{sec:learning} are trained on four cities chosen to
span diverse urban forms, Madrid, Los Angeles, Chongqing and S\~ao Paulo, from old
compact centers to sprawling modern layouts, on scenarios drawn at random from the
parameter ranges of Table~\ref{tab:params}, the \emph{base} scenario.  Evaluation, by
contrast, freezes a single scenario per city, demanding but solvable, fixed by the
procedure of Section~\ref{sec:benchmark}, on which we test the controllers over the
four training cities and, zero-shot, over sixteen more cities across five continents
(Fig.~\ref{fig:worldmap_v2}).  We then re-evaluate the same cities under the \emph{high-capacity large-van regime}, an
out-of-distribution yet realistic configuration with
larger-battery vehicles, a higher initial charge and a growing fleet-to-charger
density, to see how the controllers transfer beyond the conditions they were trained on
(Section~\ref{sec:highcap_scenario}).  The underlying question
is what a learned controller can add in a setting like this, where only a subset of the
fleet needs to charge yet those that do may face sharp contention for the chargers.

\subsection{The base scenario}
\label{sec:base_scenario}

\subsubsection{Scenario setup}
\label{sec:base_setup}

The \emph{base} scenario of Table~\ref{tab:params} instantiates the metropolitan
delivery setting of this paper, on routes longer and denser than light last-mile parcel
work, with hardware and fleet parameters common to commercial delivery today: the
two-to-six connectors per station and the
22/50/150~kW power tiers are standard charging hardware,\footnote{Per-station
connector counts and power ratings follow industry deployment reports, e.g.\
\url{https://www.gridx.ai/resources/european-ev-charging-report-2025}.} and the
fleet sizes, route lengths and time budgets match those of urban delivery shifts.

The fleet itself is parameterized after commercial delivery vans already in
large-scale service.  Battery capacity $C\sim\mathcal{U}(45,70)$~kWh spans the commercial vans that dominate
the segment.  Its lower end matches the $45$~kWh Renault Kangoo E-Tech, of which La
Poste runs some $7{,}000$~\citep{Renault2023kangoolaposte, GreenNCAP2023kangoo}, and
the $60$~kWh Mercedes eVito that Amazon fields by the thousand for urban
delivery~\citep{Amazon2020evito, EVDatabase2024evito}.  Its upper end matches the
$68$~kWh Ford E-Transit, run in the thousands by Deutsche Post/DHL ($4{,}900$~Ford~Pro
e-vans in its German parcel unit)~\citep{DHL2025etransit, FordPro2022etransit}, and the
$69$~kWh Peugeot e-Expert~\citep{EVDatabase2021eexpert}.
Energy use $e\sim\mathcal{U}(0.22,0.35)$~kWh/km follows the real-world figures
of \citet{Fiori2018energy}, whose five electric freight vans in central Rome averaged
$0.38$~kWh/km gross, or $0.28$ net of the $24.9\%$ recovered by regenerative braking,
with per-vehicle net consumption of $0.23$--$0.43$~kWh/km rising with payload (their
Table~3).  Initial charge
$\text{SoC}^0\sim\mathcal{U}(0.60,1.00)$ reflects overnight depot
charging~\citep{IEA2025gevo}, which does not always leave every van full, as observed in real EV fleet operation~\citep{Sorensen2024}.

\begin{table*}[t!]
\centering
\scriptsize
\renewcommand{\arraystretch}{1.14}
\setlength{\tabcolsep}{5pt}
\caption{Parameters of the study: the base scenario (left) and the learners
(right, NEAT, PPO with stable-baselines3, and the SA-tuned rule; unlisted PPO
values are library defaults).  Consumption rate from~\citet{Fiori2018energy}.
Scenario-level quantities are drawn per episode during training
(Section~\ref{TRAINING}); the $\varphi$ range $0.15$--$0.50$ is the training draw, while in the
20-city benchmark each city freezes one scenario draw, with $\varphi$ set to a round fraction from
$\{0.15,0.20,0.25,0.30,0.35\}$ and $\lambda_{\text{ext}}$ held at its calibrated value of 2.0
(Section~\ref{sec:benchmark}).}
\label{tab:params}\label{tab:hparams}
\begin{tabular}[t]{@{}l l r@{}}
\toprule
\multicolumn{3}{@{}l}{\textbf{Base scenario}} \\
\midrule
\multicolumn{3}{@{}l}{\textit{Scenario level} (per episode / frozen per city)} \\
$M$                       & Charging stations               & $\{3, 5, 7, 10\}$ \\
$d_M$                     & Fleet density $d_M{=}N/M$ (EVs/station) & $\mathcal{U}(5,15)$ \\
$\varphi$                 & Shift-reserve fraction          & 0.15--0.50      \\
$\lambda_{\text{ext}}$    & External arrivals (veh/h/st.)   & $\mathcal{U}(0,2)$ \\
$E_j$                     & EVSEs/station                   & $\mathcal{U}\{2,\dots,6\}$ \\
$P^{\text{evse}}_j$       & EVSE power                      & $\{22, 50, 150\}$~kW \\
\midrule
\multicolumn{3}{@{}l}{\textit{Route generation} (per route)} \\
$h_i$                     & Time budget                     & $\mathcal{U}(6, 12)$~h \\
$p_{\text{jump}}$         & Long-jump probability           & $\mathcal{U}(0.02, 0.60)$  \\
$r_L$                     & Long-jump radius                & ${\approx}\,4$--$22$~km \\
$r_S$                     & Short-jump radius (fixed)       & ${\approx}\,2$~km \\
$\bar{t}_{\text{svc}}$    & Service time (mean)             & $0.12\,\mathcal{U}(0.5,1.5)$~h \\
\midrule
\multicolumn{3}{@{}l}{\textit{Fleet and physics} (all experiments)} \\
$C$                       & Battery capacity                & $\mathcal{U}(45, 70)$~kWh \\
$e$                       & Consumption rate (kWh/km)       & $\mathcal{U}(0.22, 0.35)$ \\
$P^{\text{bms}}$          & Vehicle BMS power cap           & $\{50,80,100,150\}$~kW \\
$\text{SoC}^0$            & Initial SoC                     & $\mathcal{U}(0.60, 1.00)$ \\
$\text{SoC}_{\min}$       & Critical SoC threshold          & 0.10            \\
$H$                       & Time-budget cap                 & 12~h            \\
$K_{\text{ChS}}$          & Candidate stations/decision     & 5               \\
$\Delta_{\text{SoC}}$     & Target-SoC quantization         & 0.05            \\
$\mu_{\text{ext}}^{-1}$   & Mean ext.\ charge time          & 30~min          \\
$\Delta t_{\text{slot}}$  & Slot event interval             & 5~min           \\
$J$                       & Episodes averaged (train/eval)  & 10 / 1 \\
\bottomrule
\end{tabular}
\hfill
\begin{tabular}[t]{@{}l l r@{}}
\toprule
\multicolumn{3}{@{}l}{\textbf{Learning parameters}} \\
\midrule
\multicolumn{3}{@{}l}{\textit{NEAT}} \\
$G$                       & Population (24-core machines)   & 96 \\
$S$                       & Species (kept diverse)          & 6--12 \\
                          & Node activation (fixed)         & sigmoid \\
$I$                       & Fitness runs/genome (CRN)       & 30 \\
$n_{\text{gen}}$          & Generations (budget)            & $\le 1000$ \\
\midrule
\multicolumn{3}{@{}l}{\textit{PPO}} \\
                          & Actor/critic MLP (Tanh)         & $2{\times}64$ \\
$\epsilon$                & Clip range                      & $0.2$ \\
$\eta$                    & Learning rate                   & $2{\times}10^{-4}$ \\
$\gamma$                  & Discount                        & $0.99$ \\
$\lambda$                 & GAE parameter                   & $0.95$ \\
$T_{\text{roll}}$         & Rollout length (steps)          & $2048$ \\
$B$                       & Mini-batch                      & $128$ \\
$n_{\text{epoch}}$        & Epochs per round                & $10$ \\
$n_{\text{env}}$          & Parallel environments           & 16--24 \\
$K_{\text{frozen}}$       & Historical-pool refresh         & 20 \\
\midrule
\multicolumn{3}{@{}l}{\textit{SA-tuned rule} (search ranges)} \\
$\theta_{\text{SoC}}$     & SoC trigger                     & $[0.05, 0.50]$ \\
$\theta_{\text{max}}$     & Max target SoC                  & $[0.60, 1.00]$ \\
$\theta_{\text{margin}}$  & Safety margin                   & $[0.00, 0.25]$ \\
\bottomrule
\end{tabular}
\end{table*}

\newlength{\figvcbht}%
\begin{figure}[tb]
\centering
\begin{minipage}{0.78\linewidth}
\centering
\settoheight{\figvcbht}{\includegraphics[width=0.42\linewidth]{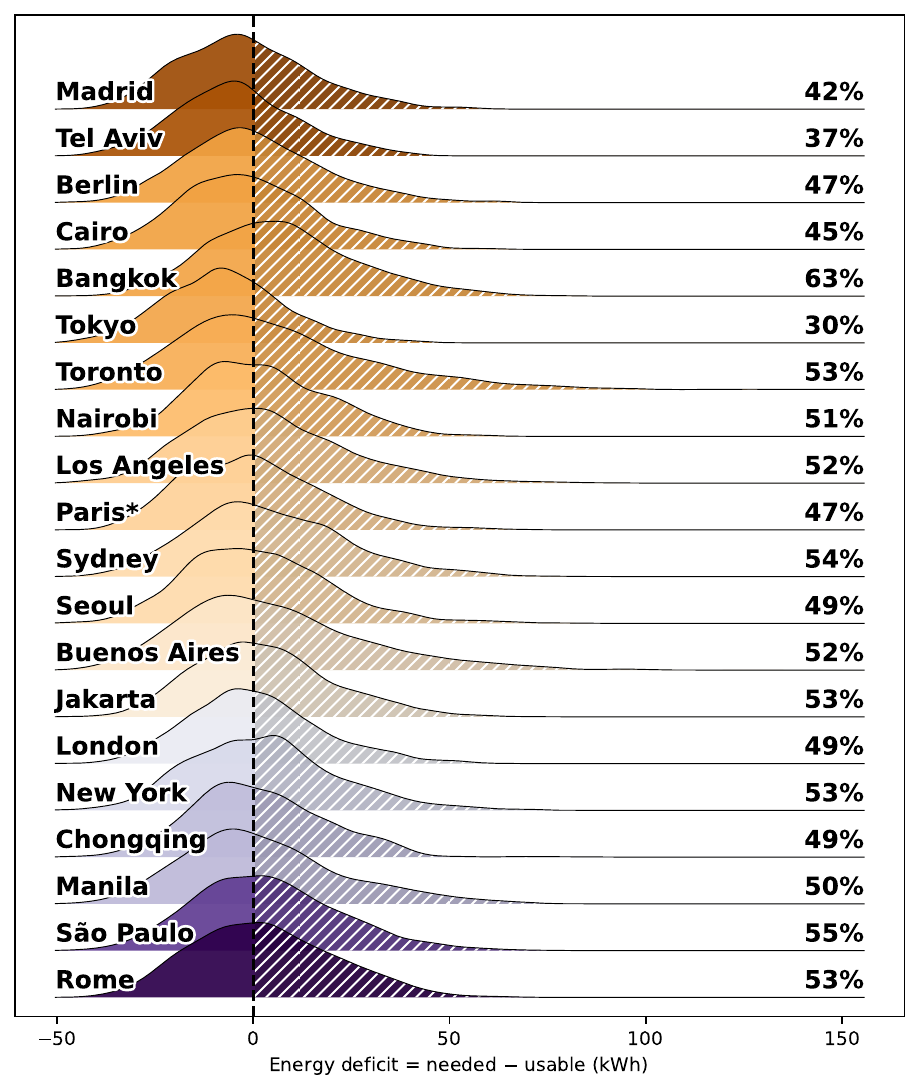}}%
\begin{subfigure}[t]{0.42\linewidth}
\centering
\includegraphics[width=\linewidth]{fig5_panel_B.pdf}
\caption{Base regime ($\text{SoC}^0\!\sim\!\mathcal{U}(0.60,1.00)$):
$50.7\%$ need a top-up}
\label{fig:operating_point_a}
\end{subfigure}\hfill
\begin{subfigure}[t]{0.42\linewidth}
\centering
\includegraphics[width=\linewidth]{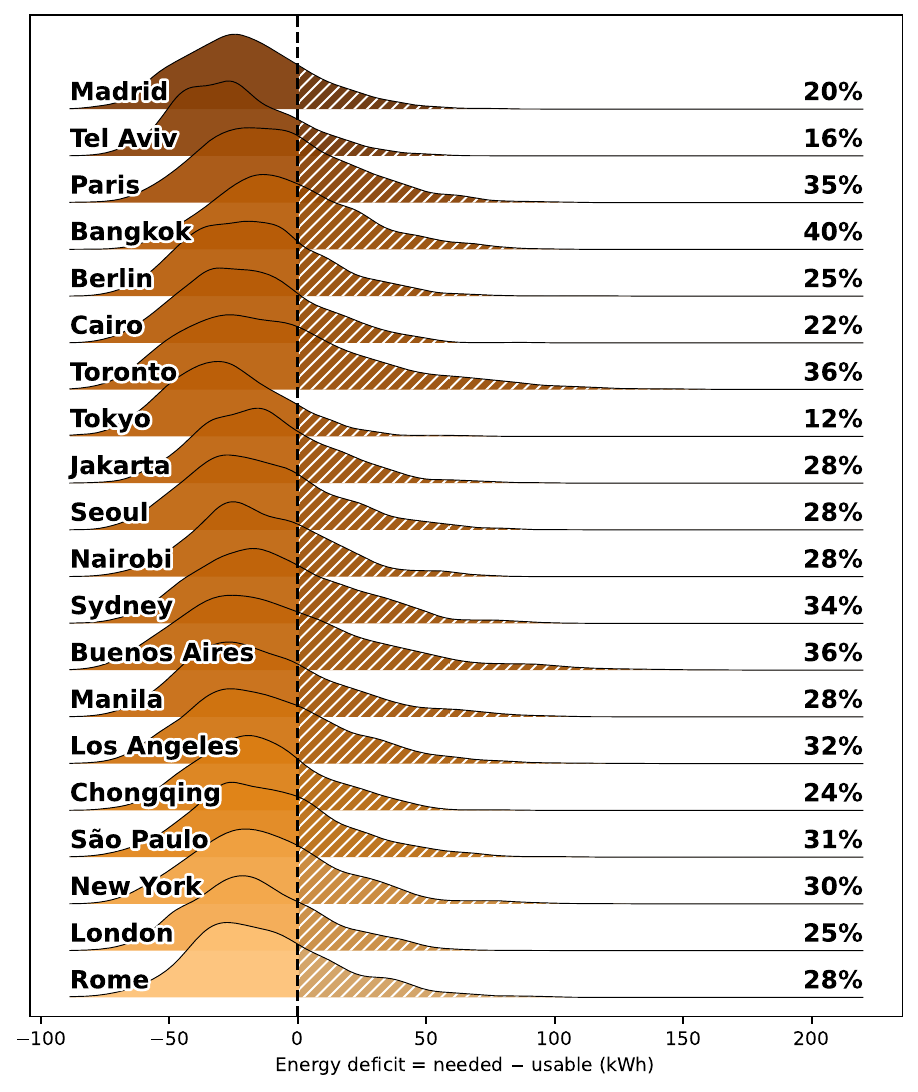}
\caption{High-capacity regime ($d_M$=25 EV/station):
$29.7\%$ need a top-up}
\label{fig:operating_point_b}
\end{subfigure}\hfill
\begin{subfigure}[t]{0.14\linewidth}
\centering
\includegraphics[height=\figvcbht]{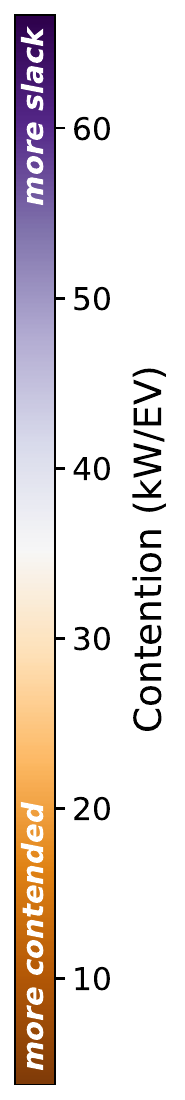}
\end{subfigure}
\end{minipage}
\caption{Per-city distribution of the per-vehicle \emph{energy deficit} over
\emph{identical} routes. Each ridge is one benchmark city; the horizontal axis is
the deficit in kWh (a shift's route energy minus the charge usable from the initial
state), and the dashed line marks zero: hatched mass to its right is the
\emph{must-charge} share, whose route exceeds its starting autonomy. Color encodes
\emph{charger contention} (kW/EV; orange $=$ more contended, low kW/EV, purple $=$ less contended, scale shared by both
panels). (a)~Base regime: $50.7\%$ must-charge. (b)~High-capacity
regime over the same routes: $29.7\%$.}
\label{fig:operating_point}
\end{figure}

With
$45$--$70$~kWh drawn at $0.22$--$0.35$~kWh/km, net of the $\text{SoC}_{\min}$=0.10
reserve, a full charge covers roughly $140$--$235$~km (median ${\sim}180$~km) under
delivery conditions, in contrast to the $256$--$330$~km these vans are rated for under the
Worldwide Harmonised Light Vehicles Test Procedure (WLTP), since delivery duty (full
payload, frequent stops) is more energy-intensive than the WLTP cycle.  Fleets seldom
start full, though: the usable range falls to a median of
about $140$~km ($p_{10}$--$p_{90}$ of $100$--$195$~km).  Delivery routes reach well
into that band and often past it.  Fig.~\ref{fig:operating_point} plots the resulting
per-vehicle \emph{energy deficit}, a shift's route energy minus what the starting charge
can supply above the reserve.  A scenario average of $50.7\%$ of the fleet falls on the
must-charge side.

Range is only one axis of difficulty.  Chargers are shared, and the fleet competes
for them with other users (Section~\ref{sec:congestion}), so a route that fits within
one charge can still time out when every connector is busy.
Fig.~\ref{fig:operating_point} colors each city by the charging power available per
EV, a proxy for scenario contention.  A third axis is the time reserve $\varphi$
(Table~\ref{tab:city_scenarios_phigrid}): the fraction of the shift held back for detours,
charging and delays, so a low $\varphi$ ($0.15$) leaves little slack to absorb a charging stop
and a high $\varphi$ ($0.35$) is forgiving.  Demanding but solvable by design
(Section~\ref{sec:benchmark}), the scenarios are hard along different axes: Bangkok carries the
heaviest energy deficit ($63\%$ must-charge), the $\varphi$=0.15 cities (Berlin, New York among
them) the tightest reserve, and Tel Aviv the sharpest contention, whereas Tokyo pairs the lowest
deficit ($30\%$) with ample slack.

\subsubsection{Benchmark scenarios}
\label{sec:benchmark}

Each benchmark scenario is frozen bit-for-bit under an isolated seed (its station
subset, per-station EVSE count and power, fleet size and reserve fraction) and held
out from training.  The scenarios are sized to be demanding but solvable: we draw them
at random from the operating distribution and keep those the Oracle resolves in the
$99$--$100\%$ band, so each instance admits an almost-complete solution with no slack
to spare, dimensioned to a service-level target as in real fleet and
charging-infrastructure planning \citep{Varma2026fleet, Yang2023sizing}.  The
20-city benchmark set is diverse in station count, fleet size and reserve
fraction, and Table~\ref{tab:city_scenarios_phigrid} gives each city's frozen design
and the resulting route statistics, tying each deployment to
Fig.~\ref{fig:operating_point}.

\begin{table*}[t]
\centering
\caption{The 20-city benchmark, frozen per city.  \emph{Design}: stations $M$, fleet $N$, density $d_M=N/M$, reserve $\varphi$, mean EVSE count $\bar E$ and power $P^{\text{evse}}$ (kW).  \emph{Operation} (per-EV real routes, cf.\ Fig.~\ref{fig:operating_point}): route length (km) and demanded energy (kWh) as mean and $p_{90}$, charger contention (kW/EV), and must-charge fraction (\%).  Paris uses a wider $25$\,km map, Tel Aviv the hardest-corner setup.}
\label{tab:city_scenarios_phigrid}
\scriptsize\setlength{\tabcolsep}{3pt}\renewcommand{\arraystretch}{1.05}
\begin{tabular}{l r r r r r r r r r r r r}
\toprule
 & \multicolumn{6}{c}{Design} & \multicolumn{2}{c}{Route (km)} & \multicolumn{2}{c}{Needed (kWh)} & Contention & Must-charge \\
\cmidrule(lr){2-7}\cmidrule(lr){8-9}\cmidrule(lr){10-11}
City & $M$ & $N$ & $d_M$ & $\varphi$ & $\bar E$ & $P^{\text{evse}}$ & mean & $p_{90}$ & mean & $p_{90}$ & (kW/EV) & (\%) \\
\midrule
\multicolumn{13}{l}{\textit{Training cities}} \\
Madrid, Spain & $3$ & $37$ & $12.3$ & $0.25$ & $3.0$ & $31$ & $135$ & $197$ & $38$ & $59$ & $7.6$ & $42$ \\
Los Angeles, USA & $10$ & $138$ & $13.8$ & $0.30$ & $3.9$ & $92$ & $153$ & $235$ & $44$ & $69$ & $25.9$ & $52$ \\
Chongqing, China & $3$ & $26$ & $8.7$ & $0.35$ & $3.7$ & $107$ & $146$ & $209$ & $41$ & $61$ & $45.4$ & $49$ \\
S\~ao Paulo, Brazil & $10$ & $71$ & $7.1$ & $0.20$ & $4.4$ & $94$ & $154$ & $227$ & $44$ & $65$ & $58.5$ & $55$ \\
\midrule
\multicolumn{13}{l}{\textit{Hold-out cities}} \\
Bangkok, Thailand & $10$ & $132$ & $13.2$ & $0.25$ & $3.9$ & $69$ & $168$ & $243$ & $48$ & $71$ & $20.3$ & $63$ \\
Berlin, Germany & $10$ & $135$ & $13.5$ & $0.15$ & $3.2$ & $84$ & $143$ & $213$ & $41$ & $62$ & $20.0$ & $47$ \\
Buenos Aires, Argentina & $7$ & $69$ & $9.9$ & $0.35$ & $3.6$ & $85$ & $161$ & $266$ & $46$ & $76$ & $30.7$ & $52$ \\
Cairo, Egypt & $5$ & $67$ & $13.4$ & $0.35$ & $4.6$ & $59$ & $138$ & $210$ & $39$ & $60$ & $20.2$ & $45$ \\
Jakarta, Indonesia & $10$ & $90$ & $9.0$ & $0.15$ & $4.2$ & $69$ & $150$ & $217$ & $43$ & $63$ & $32.1$ & $53$ \\
London, UK & $7$ & $97$ & $13.9$ & $0.15$ & $3.9$ & $136$ & $143$ & $209$ & $41$ & $61$ & $37.8$ & $49$ \\
Manila, Philippines & $3$ & $21$ & $7.0$ & $0.15$ & $4.3$ & $74$ & $150$ & $236$ & $43$ & $68$ & $45.8$ & $50$ \\
Nairobi, Kenya & $3$ & $38$ & $12.7$ & $0.30$ & $4.0$ & $74$ & $149$ & $213$ & $42$ & $62$ & $23.4$ & $51$ \\
New York, USA & $7$ & $85$ & $12.1$ & $0.15$ & $4.1$ & $121$ & $152$ & $226$ & $43$ & $64$ & $41.4$ & $53$ \\
Paris, France & $10$ & $77$ & $7.7$ & $0.20$ & $4.1$ & $50$ & $161$ & $235$ & $46$ & $68$ & $26.6$ & $47$ \\
Rome, Italy & $3$ & $27$ & $9.0$ & $0.25$ & $4.0$ & $150$ & $150$ & $224$ & $43$ & $65$ & $66.7$ & $53$ \\
Seoul, South Korea & $10$ & $104$ & $10.4$ & $0.15$ & $3.8$ & $77$ & $147$ & $224$ & $42$ & $63$ & $28.2$ & $49$ \\
Sydney, Australia & $10$ & $107$ & $10.7$ & $0.30$ & $4.2$ & $72$ & $157$ & $236$ & $45$ & $69$ & $28.1$ & $54$ \\
Tel Aviv, Israel & $3$ & $45$ & $15.0$ & $0.15$ & $3.3$ & $41$ & $128$ & $189$ & $36$ & $55$ & $9.0$ & $37$ \\
Tokyo, Japan & $3$ & $41$ & $13.7$ & $0.30$ & $4.3$ & $65$ & $119$ & $179$ & $33$ & $51$ & $20.5$ & $30$ \\
Toronto, Canada & $10$ & $127$ & $12.7$ & $0.35$ & $4.0$ & $69$ & $162$ & $270$ & $46$ & $78$ & $21.7$ & $53$ \\
\bottomrule
\end{tabular}
\end{table*}

\subsubsection{Controllers training}
\label{TRAINING}

The four learned controllers, two NEAT and two PPO, each come from a single
training run, from scratch by domain randomization, a fresh scenario per
episode~\citep{Tobin2017domainrand}, over the four
base-scenario cities, and are deployed zero-shot.  Every episode draws a fresh
scenario from one of the four training cities, sampled from the ranges of
Table~\ref{tab:params}.  Their hyperparameters are in Table~\ref{tab:hparams}.

\begin{figure*}[!tbp]
\centering
\includegraphics[width=0.85\textwidth]{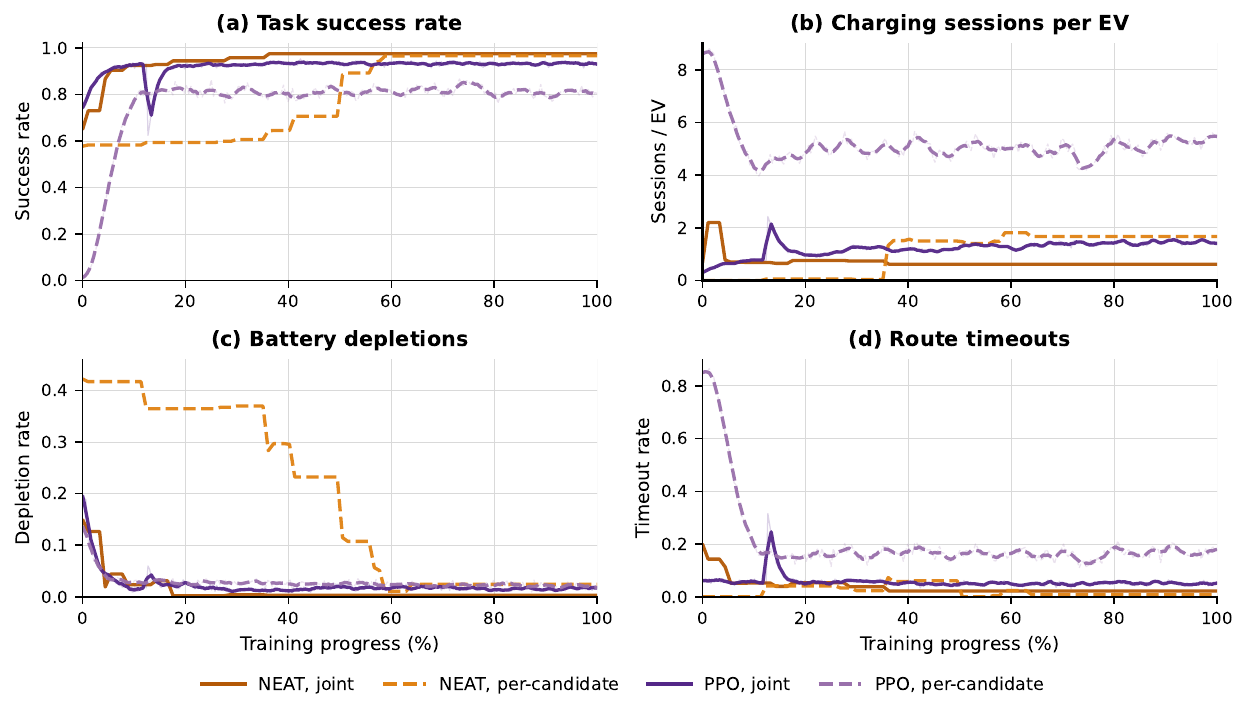}
\caption{Training dynamics of the four learned controllers.  (a)~task success rate,
(b)~charging sessions per vehicle, (c)~battery depletions, (d)~route timeouts.  Progress is
normalized to $[0,100]\%$ (NEAT in generations, PPO in steps).}
\label{fig:training_generalist}
\end{figure*}

All controllers share a comparable training budget of order ${\sim}300$~CPU-h.
NEAT trains from scratch, whereas PPO reaches the same level only with the heavier
machinery of Section~\ref{sec:ppo}: the single-agent ego-learner reduction, the
behavioral-cloning warm-start and the historical policy pool.
Fig.~\ref{fig:training_generalist} traces the training dynamics of the four learned
controllers, with the NEAT-per-candidate variant the slowest to converge.  The PPO and NEAT curves are not comparable in
absolute level: NEAT scores a homogeneous fleet running the current genome, while each PPO
curve reports the aggregate reward of a mixed fleet (one ego-learner among $N-1$ frozen
past policies, Section~\ref{sec:ppo}).  The
SA-tuned rule is fit separately by simulated annealing on the same reward
(Section~\ref{sec:sa_heuristic}), and is therefore absent from
Fig.~\ref{fig:training_generalist}.

\subsubsection{Scenario evaluation}
\label{EVALUATION}

The controllers are evaluated on a frozen benchmark of 20 cities, in
configurations never seen during training, under common random numbers (CRN),
so all controllers face bit-identical fleets, routes and congestion.  For each
frozen benchmark configuration we run up to 50 independent episodes per
city.\footnote{The Oracle is limited to 20 episodes per city because its
per-episode planning is computationally heavy.}  Each evaluation episode varies
the shift's routes and each vehicle's initial SoC, as in a real fleet, so the
per-city spread we report is route variance alone.  Training and evaluation
alike run under the per-city calibrated routing surrogate of
Section~\ref{sec:stat_model}, whose residual error, shared identically across
controllers under CRN, does not bias the comparison.

The evaluation metric is the per-fleet normalized reward $R\in[-1,1]$ of
Section~\ref{sec:reward}.  Each table entry is a controller's mean reward over
these episodes, alongside the Oracle and \emph{greedy} references, and where
informative we also report route-completion (\texttt{ok}), timeout and
battery-depletion rates, the number of charging sessions per vehicle, and the best
controller's gap to the Oracle (the \emph{gap} column of
Table~\ref{tab:benchmark_phigrid}).

\subsection{Results on the base benchmark}
\label{sec:res_base}

\begin{table*}[t]
\centering
\caption{Benchmark: mean per-fleet reward $\bar R\in[-1,1]$ (Section~\ref{sec:reward}) per city, one frozen scenario each; all learned controllers are generalist zero-shot, best reactive per row highlighted (purple: PPO, orange: NEAT).  \emph{Oracle}: best simulated-annealing plan.  \emph{Gap}: Oracle minus the best reactive controller in that city (mean $0.009$).  \emph{per-cand}/\emph{joint}: per-candidate/joint policy.}
\label{tab:benchmark_phigrid}
\scriptsize\setlength{\tabcolsep}{2.2pt}\renewcommand{\arraystretch}{1.0}
\begin{tabular}{l cc ccc cc cc}
\toprule
& & & & & & \multicolumn{2}{c}{NEAT} & \multicolumn{2}{c}{PPO} \\
\cmidrule(lr){7-8}\cmidrule(lr){9-10}
City & Oracle & Gap & \emph{greedy} & \emph{null} & SA & per-cand & joint & per-cand & joint \\
\midrule
\multicolumn{10}{l}{\textit{Training cities}} \\
Madrid, Spain & $+1.000$ & $+0.020$ & $+0.123$ & $+0.172$ & $+0.753$ & $+0.799$ & $+0.519$ & \cellcolor{ppocell}$\mathbf{+0.980}$ & $+0.947$ \\
Los Angeles, USA & $+0.979$ & $+0.004$ & $+0.430$ & $-0.036$ & $+0.791$ & $+0.879$ & $+0.963$ & $+0.971$ & \cellcolor{ppocell}$\mathbf{+0.975}$ \\
Chongqing, China & $+0.985$ & $+0.012$ & $+0.466$ & $+0.025$ & $+0.649$ & \cellcolor{neatcell}$\mathbf{+0.973}$ & $+0.951$ & $+0.966$ & $+0.966$ \\
S\~ao Paulo, Brazil & $+0.985$ & $+0.008$ & $+0.488$ & $-0.090$ & $+0.801$ & $+0.927$ & $+0.899$ & \cellcolor{ppocell}$\mathbf{+0.977}$ & $+0.975$ \\
\midrule
\multicolumn{10}{l}{\textit{Hold-out cities}} \\
Bangkok, Thailand & $+0.981$ & $+0.001$ & $+0.258$ & $-0.239$ & $+0.750$ & $+0.888$ & $+0.932$ & \cellcolor{ppocell}$\mathbf{+0.980}$ & $+0.976$ \\
Berlin, Germany & $+0.987$ & $+0.004$ & $+0.442$ & $+0.072$ & $+0.844$ & $+0.804$ & $+0.930$ & \cellcolor{ppocell}$\mathbf{+0.983}$ & $+0.976$ \\
Buenos Aires, Argentina & $+0.996$ & $+0.018$ & $+0.407$ & $-0.031$ & $+0.875$ & $+0.532$ & $+0.962$ & $+0.911$ & \cellcolor{ppocell}$\mathbf{+0.978}$ \\
Cairo, Egypt & $+0.988$ & $+0.007$ & $+0.571$ & $+0.117$ & $+0.774$ & $+0.931$ & $+0.976$ & $+0.972$ & \cellcolor{ppocell}$\mathbf{+0.981}$ \\
Jakarta, Indonesia & $+0.981$ & $+0.002$ & $+0.303$ & $-0.041$ & $+0.768$ & $+0.897$ & $+0.878$ & \cellcolor{ppocell}$\mathbf{+0.979}$ & $+0.966$ \\
London, UK & $+0.982$ & $+0.001$ & $+0.640$ & $+0.044$ & $+0.889$ & $+0.933$ & $+0.937$ & \cellcolor{ppocell}$\mathbf{+0.981}$ & $+0.960$ \\
Manila, Philippines & $+1.000$ & $+0.018$ & $+0.447$ & $+0.034$ & $+0.932$ & $-0.434$ & \cellcolor{neatcell}$\mathbf{+0.982}$ & $+0.898$ & $+0.973$ \\
Nairobi, Kenya & $+1.000$ & $+0.019$ & $+0.358$ & $-0.017$ & $+0.872$ & $+0.937$ & $+0.958$ & \cellcolor{ppocell}$\mathbf{+0.981}$ & $+0.967$ \\
New York, USA & $+0.990$ & $+0.009$ & $+0.504$ & $-0.045$ & $+0.923$ & $+0.882$ & $+0.950$ & \cellcolor{ppocell}$\mathbf{+0.981}$ & $+0.970$ \\
Paris, France & $+0.985$ & $+0.003$ & $+0.660$ & $+0.079$ & $+0.866$ & $+0.936$ & $+0.964$ & \cellcolor{ppocell}$\mathbf{+0.982}$ & $+0.976$ \\
Rome, Italy & $+1.000$ & $+0.017$ & $+0.660$ & $-0.052$ & $+0.843$ & $+0.961$ & \cellcolor{neatcell}$\mathbf{+0.983}$ & \cellcolor{ppocell}$\mathbf{+0.983}$ & $+0.961$ \\
Seoul, South Korea & $+0.991$ & $+0.008$ & $+0.603$ & $+0.029$ & $+0.952$ & $+0.821$ & $+0.961$ & \cellcolor{ppocell}$\mathbf{+0.983}$ & $+0.968$ \\
Sydney, Australia & $+0.987$ & $+0.004$ & $+0.520$ & $-0.071$ & $+0.798$ & $+0.901$ & $+0.970$ & \cellcolor{ppocell}$\mathbf{+0.983}$ & $+0.981$ \\
Tel Aviv, Israel & $+1.000$ & $+0.010$ & $+0.419$ & $+0.276$ & $+0.935$ & $+0.801$ & $+0.965$ & $+0.966$ & \cellcolor{ppocell}$\mathbf{+0.990}$ \\
Tokyo, Japan & $+0.998$ & $+0.015$ & $+0.567$ & $+0.408$ & $+0.850$ & $+0.929$ & $+0.966$ & $+0.945$ & \cellcolor{ppocell}$\mathbf{+0.983}$ \\
Toronto, Canada & $+0.985$ & $+0.006$ & $+0.640$ & $-0.043$ & $+0.964$ & $+0.457$ & \cellcolor{neatcell}$\mathbf{+0.979}$ & $+0.882$ & $+0.957$ \\
\bottomrule
\end{tabular}
\end{table*}

The base-benchmark results are collected in Table~\ref{tab:benchmark_phigrid}.  The PPO
controllers reach near-Oracle service zero-shot, within a mean gap of $0.019$ of the Oracle
($+0.990$) and never seeing the shift in advance.  Averaged over both structures they reach
$+0.968$ against NEAT's $+0.860$, best reactive in $16$ of the $20$ cities.  PPO-per-candidate
wins the most individual cities ($11$ of $20$), yet PPO-joint leads on average, at a gap of
$0.019$ against $0.026$, since per-candidate pays larger penalties where it loses.  NEAT-joint
stays competitive (gap $0.059$), whereas the evolved per-candidate one lags far behind
($+0.788$, Section~\ref{sec:highcap_scenario}).

\begin{figure*}[!tbp]
\centering
\includegraphics[width=0.85\textwidth]{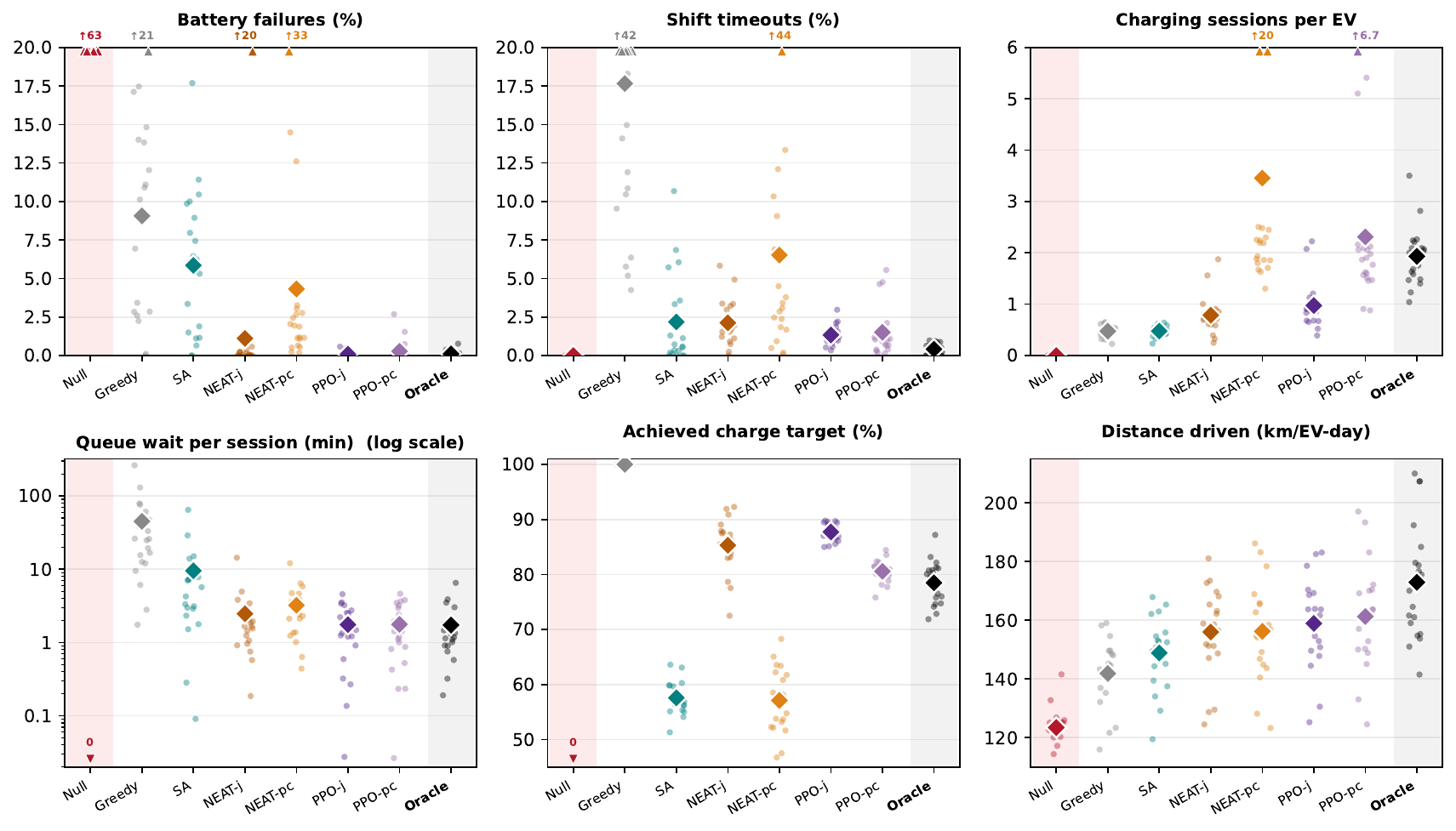}
\caption{Per-city operational signature of the controllers (\emph{null}, \emph{greedy},
the NEAT and PPO variants, and the Oracle).  Each panel shows one metric as a
per-city beeswarm, the mean marked by a diamond: battery failures (\%), shift timeouts
(\%), charging sessions per EV, queue wait per session (min, log), achieved charge target
(\%) and distance (km/EV-day).  Off-scale points appear as carets at the panel top with
their value annotated.  Full ranking in Table~\ref{tab:benchmark_phigrid}.}
\label{fig:operational}
\end{figure*}

\begin{figure*}[!tbp]
\centering
\includegraphics[width=\textwidth]{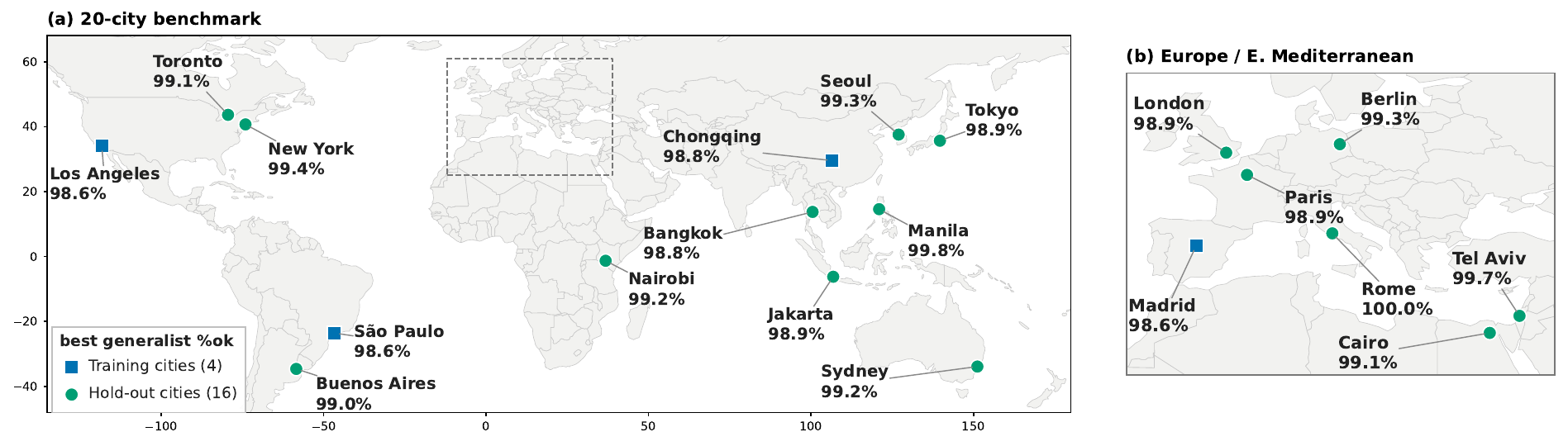}
\caption{Geographic span of the 20-city benchmark.  Blue squares: the four training
cities; green circles: the sixteen held-out cities.  Labels give each city's best zero-shot
shift-completion rate (\%ok, Table~\ref{tab:benchmark_phigrid}).}
\label{fig:worldmap_v2}
\end{figure*}

A further result worth noting is that reading the occupancy signal adds $+0.139$ of reward over the congestion-blind SA-tuned rule,
positive in all 20 cities and up to $+0.324$ in Chongqing, and cuts the per-session queue wait
to about $2$ minutes against $9.5$ for that rule and $45$ for \emph{greedy}
(Fig.~\ref{fig:operational}).  The best reactive controller in each city finishes at least
$98.6\%$ of shifts on time (Fig.~\ref{fig:worldmap_v2}), against the Oracle's $99.5\%$.  Harder
cities cost even the Oracle a little reward, and greedy and
\emph{null} degrade with them, but
PPO and NEAT-joint stay high and homogeneous across all 20.

As for the operating statistics shown in Fig.~\ref{fig:operational}, the learned controllers
charge sparingly, below a full battery (exit SoC $57$--$88\%$).  PPO-joint charges about $0.97$ times per vehicle, half the
Oracle's $1.93$ and well under PPO-per-candidate's $2.31$ and NEAT-per-candidate's $3.46$, the
latter above the Oracle itself.  NEAT-joint is similar ($0.79$), close to the threshold rules
(SA $0.48$, greedy $0.47$), which charge less only by dropping service.  The same service from
fewer charges favors battery health, something the planner has no reason to do.  The learned
controllers all cover about $156$--$161$~km per vehicle-day, though this distance is biased by
completion: a controller that does not finish drives less (null $123$, greedy $142$), so it is
not a clean measure of charging overhead.

\subsection{Policy interpretability}
\label{sec:interpretability}

Analysing the controllers' operation internally, we find that every controller starts charging as the energy margin $s_2$ runs
low, penalises the detour $s_3$, and ranks the stations before choosing, preferring a freer one
when its gain $s_4$ is higher, none of it enforced.

Comparing the two PPO structures isolates a design question: whether a candidate is better scored
from its own signals alone (per-candidate) or from all candidates at once (joint).  On the base
benchmark the two land almost together (Table~\ref{tab:benchmark_phigrid}), so the
local signals of a single candidate already carry most of what the decision needs.  The two forms diverge only when the best station depends on the others.  The joint form
sees every candidate at once and lowers a candidate's score when a rival has free capacity (a
cross-candidate sensitivity averaging about $-3$, steepening to $-6$ to $-12$ in contended states), so it passes a congested station for a farther free
one, where the per-candidate form scores each station on its own and cannot.  The
same coordination shows in how often each charges: the joint form scores the no-charge option
jointly with the candidates, so the same suppression can pull a candidate below not charging, and
it postpones, settling on fewer stops.

Under contention this difference grows.  In Manila, the per-candidate forms over-charge, $6.7$ sessions
per vehicle for PPO and $19.6$ for NEAT against PPO-joint's $1.2$, each vehicle topping up at
nearly every stop because no single-station score reveals that a neighbor is free.  For PPO this
costs only efficiency, since its network is expressive enough for the per-candidate design to work.  For the
evolved NEAT-per-candidate it is fatal: a tiny genome, three hidden nodes and a dozen
connections against PPO's two $64$-unit layers, so its scores sit in a narrow band and lose all discrimination
outside the trained range, where PPO's stay wide and stable, and its reward turns negative
($-0.434$, Table~\ref{tab:benchmark_phigrid}), the fleet lost to a mix of timeouts and
depletions.  NEAT-joint has the same input but not the behavior, its evolved network of about
eighty connections saturating until the cross-candidate coupling collapses to zero.

The per-candidate scoring network is invariant to the number of candidate stations, so the same
policy applies unchanged as the candidate set grows or shrinks.  Additionally, the joint's
cross-candidate coordination comes at the expense of this flexibility: it is tied to a fixed
number of candidate stations and must be rebuilt if that number changes.

\begin{figure*}[!tbp]
\centering
\includegraphics[width=0.72\textwidth]{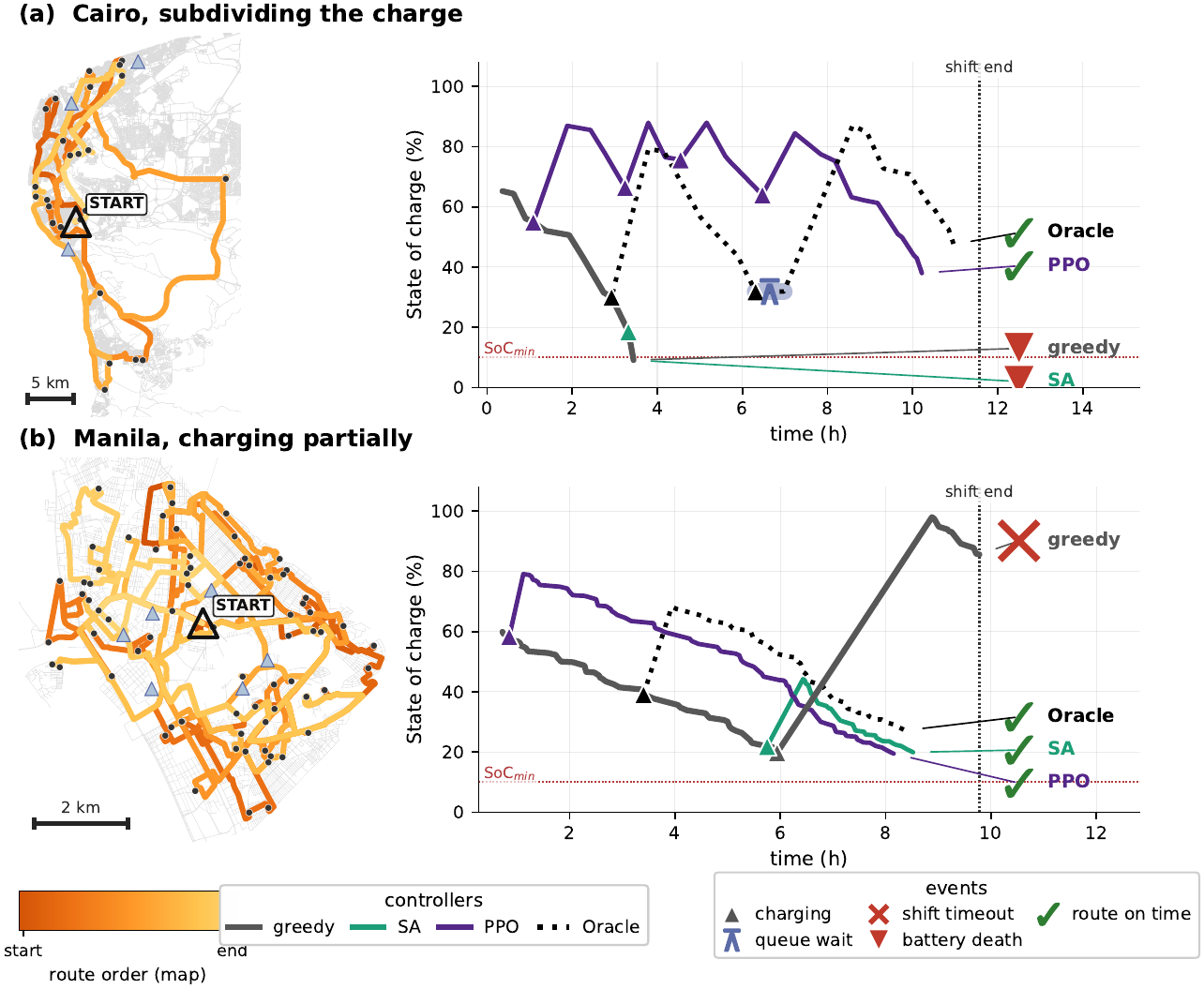}
\caption{Two single-vehicle shifts under identical route conditions across the four
controllers.  Each row: left, the shared delivery route colored by visit order (dark to
light, start to end); right, state of charge over the shift for \emph{greedy}, SA-tuned,
PPO-joint and the Oracle.  (a) Cairo, subdividing the charge; (b) Manila, charging partially.}
\label{fig:trace_mechanisms}
\label{fig:trace_subdivide}
\label{fig:trace_restraint}
\end{figure*}

\begin{figure*}[tbp]
\centering
\includegraphics[width=0.72\textwidth]{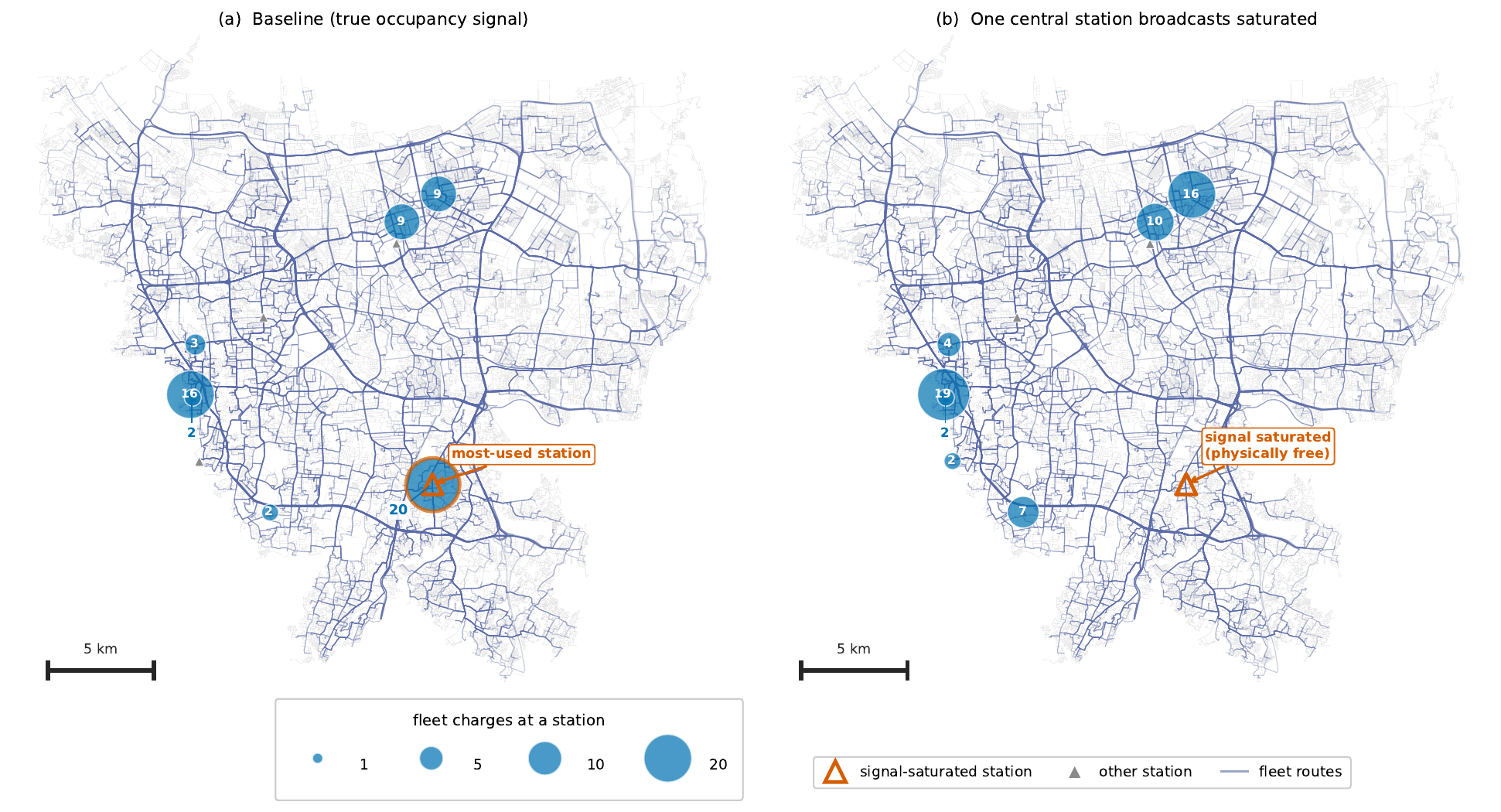}
\caption{Station avoidance from the public signal alone (Jakarta, PPO-joint).  (a) baseline;
(b) one central station broadcasts a saturated occupancy signal while staying physically free.
Marker area is the number of fleet charges at a station; the manipulated station (orange ring)
is the most used at baseline ($33\%$ of charges) and falls to zero when saturated, its load
spreading to neighbors.  Faint lines are fleet routes.}
\label{fig:saturation}
\end{figure*}

Finally, the runs show emergent coordination behaviors like those mentioned in the introduction:
\begin{itemize}
\item \textbf{Subdividing the charge.}  Fig.~\ref{fig:trace_subdivide}(a) follows one Cairo
vehicle whose route needs more than one stop: PPO-joint and the planner spread the energy over
short top-ups and keep the battery in a working band, while greedy and the SA-tuned rule
each commit to one large charge and lose the shift to that single long stop.
\item \textbf{Charging partially.}  PPO stops just past the constant-voltage knee at $0.80$,
where the last points of charge cost disproportionately more time.
Fig.~\ref{fig:trace_restraint}(b) follows a vehicle where one charge is enough: PPO-joint, the
SA-tuned rule and the Oracle each stop between $0.45$ and $0.85$ and finish, whereas
greedy charges to a full battery and the slow tail above the knee overruns the shift.
The NEAT controllers instead drive toward a full charge and stop short only when a feasibility
limit caps them.
\item \textbf{Avoiding a busy station.}  Forcing one central Jakarta station to broadcast a
saturated occupancy signal while leaving it physically free (Fig.~\ref{fig:saturation}) empties
it: its share of the fleet's charging falls from the largest of any station ($33\%$) to zero,
redistributed to its neighbors, at almost no cost (fleet reward $1.00$ to $0.98$).  The fleet
coordinates around a station it never communicates about, purely by reacting to what it
broadcasts.  This also shows the robustness of the learning agents to unforeseen infrastructure
events.
\end{itemize}

\subsection{Beyond the training regime: capacity and density}
\label{sec:highcap_scenario}

\begin{figure*}[!tbp]
\centering
\includegraphics[width=\textwidth]{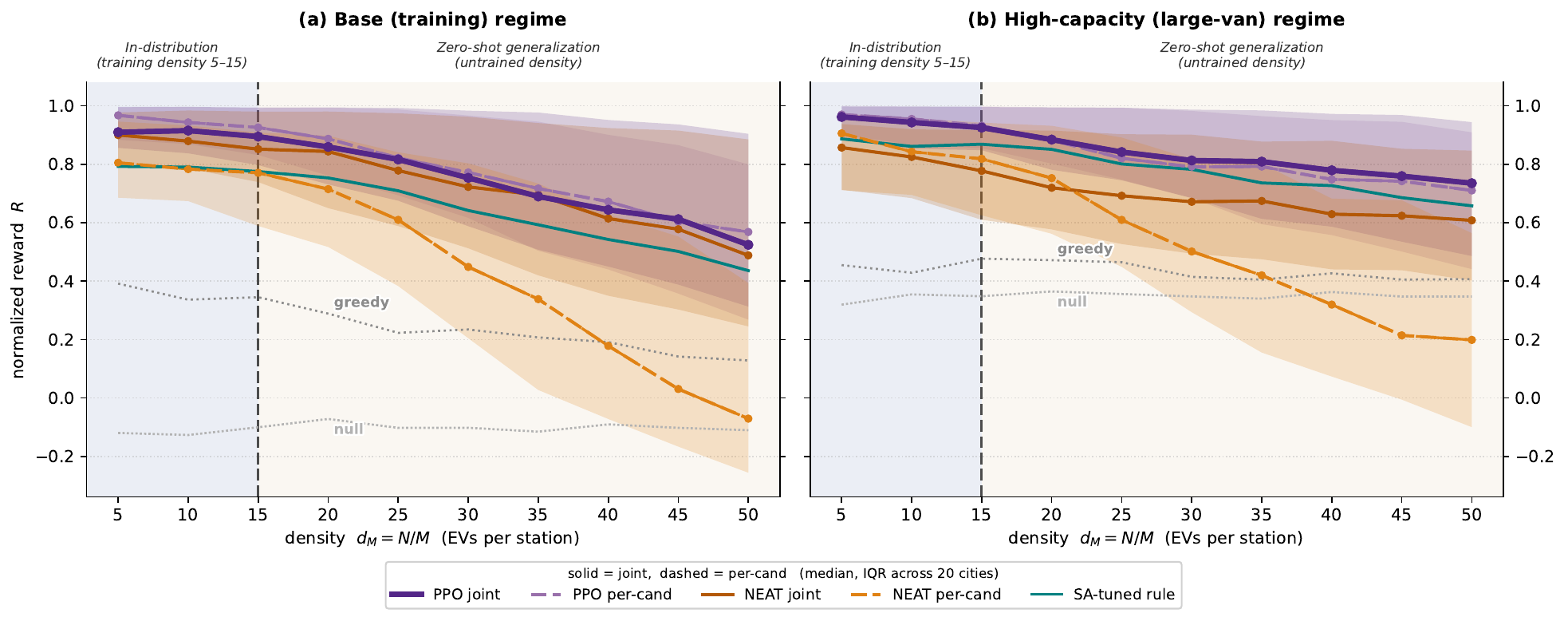}
\caption{Zero-shot density transfer: demand density $d_M$ (EVs per station) swept from
5 to 50 in (a)~the base regime and (b)~the high-capacity regime.  Each curve is
the median across the 20 cities (band: inter-city IQR).  Color: optimizer (purple PPO,
orange NEAT); line: structure (solid joint, dashed per-candidate); SA-tuned rule teal,
\emph{greedy}/\emph{null} inline.  The band $d_M\!\in\![5,15]$ is in-distribution, beyond it
is extrapolation with no re-training.  PPO, NEAT-joint and the SA-tuned rule hold in both
regimes; the evolved per-candidate controller collapses, turning negative in the base
regime.}
\label{fig:density_transfer}
\end{figure*}

In this section we investigate the controllers out of the training distribution.  For that we
choose a second scenario of interest, a high-capacity regime: the fleet is swapped for
large vans, capacity
$C\sim\mathcal{U}(90,120)$~kWh and consumption $e\sim\mathcal{U}(0.40,0.55)$~kWh/km, starting
fuller ($\mathrm{SoC}^0\sim\mathcal{U}(0.80,1.00)$), at a raised demand density of 25 EVs per
station and a reserve $\varphi$=0.20, keeping the base benchmark's road networks and station layouts.  These
capacities match large delivery vans already in service, the $113$~kWh Mercedes eSprinter that
Amazon fields for urban delivery and the $110$~kWh Fiat E-Ducato.  The regime is out of
distribution yet operationally realistic: the larger batteries carry more range, so fewer vans
need a mid-shift top-up, $29.7\%$ against the base $50.7\%$ (Fig.~\ref{fig:operating_point_b}),
but those that do meet sharper contention for the same chargers.  We also sweep the demand
density from 5 to 50 EVs per station, well past the training range
(Fig.~\ref{fig:density_transfer}).

Deployed zero-shot at density 25, the controllers keep the ordering of the base benchmark
(Table~\ref{tab:offdist_transfer}): the PPO controllers transfer with least loss, the
SA-tuned rule stays close behind, NEAT-joint holds, and the evolved NEAT-per-candidate
controller loses the most.  The density sweep shows the same ordering reproduces across every
density (Fig.~\ref{fig:density_transfer}), the high-capacity regime the more robust of the two,
since larger batteries recharge less often and the fixed infrastructure absorbs the added demand
with more margin.  Only the evolved NEAT-per-candidate breaks the pattern.  Scoring one candidate at a time, its
scores flip outside the trained range once travel times fall far below it, so it charges on
nearly every decision and the fleet times out or depletes, turning the reward negative.  The joint controller
and PPO's own per-candidate variant do not, so the collapse is specific to its tiny genome, not
to the per-candidate structure.  Another effect is that PPO-joint charges less here than in the base scenario, about $0.51$
sessions per vehicle, as expected from the larger energy headroom of these EVs.

\begin{table*}[t]
\centering
\caption{Zero-shot transfer to the high-capacity \emph{off-distribution} regime (large vans $C\!\sim\!\mathcal{U}(90,120)$, $e\!\sim\!\mathcal{U}(0.40,0.55)$, $\mathrm{SoC}^0\!\sim\!\mathcal{U}(0.80,1.00)$, density $25$ EVs/station, $\varphi$=0.20), never seen at training and no re-training.  Mean reward $\bar R\in[-1,1]$ per city, best reactive highlighted (purple: PPO, orange: NEAT, green: SA).  PPO transfers best, SA stays robust, NEAT (especially per-candidate) degrades most.}
\label{tab:offdist_transfer}
\scriptsize\setlength{\tabcolsep}{2.2pt}\renewcommand{\arraystretch}{1.0}
\begin{tabular}{l r ccc cc cc}
\toprule
& & & & & \multicolumn{2}{c}{NEAT} & \multicolumn{2}{c}{PPO} \\
\cmidrule(lr){6-7}\cmidrule(lr){8-9}
City & $N$ & \emph{null} & \emph{greedy} & SA & per-cand & joint & per-cand & joint \\
\midrule
\multicolumn{9}{l}{\textit{Training cities}} \\
Madrid, Spain & 75 & $+0.542$ & $+0.548$ & $+0.633$ & $+0.188$ & $+0.552$ & $+0.814$ & \cellcolor{ppocell}$\mathbf{+0.838}$ \\
Los Angeles, USA & 250 & $+0.162$ & $+0.246$ & $+0.634$ & $+0.568$ & $+0.633$ & $+0.762$ & \cellcolor{ppocell}$\mathbf{+0.770}$ \\
Chongqing, China & 75 & $-0.020$ & $+0.046$ & $+0.298$ & $+0.506$ & $+0.428$ & $+0.567$ & \cellcolor{ppocell}$\mathbf{+0.578}$ \\
S\~ao Paulo, Brazil & 250 & $+0.416$ & $+0.470$ & $+0.811$ & $+0.935$ & $+0.842$ & \cellcolor{ppocell}$\mathbf{+0.979}$ & $+0.976$ \\
\midrule
\multicolumn{9}{l}{\textit{Hold-out cities}} \\
Bangkok, Thailand & 250 & $+0.076$ & $+0.215$ & $+0.600$ & $+0.586$ & $+0.627$ & $+0.821$ & \cellcolor{ppocell}$\mathbf{+0.830}$ \\
Berlin, Germany & 250 & $+0.600$ & $+0.670$ & $+0.925$ & $+0.831$ & $+0.951$ & $+0.993$ & \cellcolor{ppocell}$\mathbf{+0.997}$ \\
Buenos Aires, Argentina & 175 & $+0.008$ & $+0.253$ & $+0.362$ & $-0.258$ & \cellcolor{neatcell}$\mathbf{+0.529}$ & $+0.469$ & $+0.508$ \\
Cairo, Egypt & 125 & $+0.204$ & $+0.148$ & \cellcolor{bestcell}$\mathbf{+0.706}$ & $+0.370$ & $+0.407$ & $+0.564$ & $+0.592$ \\
Jakarta, Indonesia & 250 & $+0.539$ & $+0.592$ & $+0.917$ & $+0.878$ & $+0.895$ & \cellcolor{ppocell}$\mathbf{+0.997}$ & $+0.994$ \\
London, UK & 175 & $+0.623$ & $+0.767$ & $+0.971$ & $+0.982$ & $+0.987$ & $+0.996$ & \cellcolor{ppocell}$\mathbf{+0.997}$ \\
Manila, Philippines & 75 & $+0.528$ & $+0.536$ & $+0.907$ & $+0.210$ & $+0.927$ & $+0.924$ & \cellcolor{ppocell}$\mathbf{+0.996}$ \\
Nairobi, Kenya & 75 & $+0.188$ & $+0.249$ & $+0.646$ & $+0.520$ & $+0.485$ & $+0.776$ & \cellcolor{ppocell}$\mathbf{+0.787}$ \\
New York, USA & 175 & $+0.473$ & $+0.602$ & $+0.955$ & $+0.954$ & $+0.965$ & \cellcolor{ppocell}$\mathbf{+0.993}$ & \cellcolor{ppocell}$\mathbf{+0.993}$ \\
Paris, France & 250 & $+0.295$ & $+0.288$ & \cellcolor{bestcell}$\mathbf{+0.829}$ & $+0.633$ & $+0.686$ & $+0.819$ & $+0.826$ \\
Rome, Italy & 75 & $+0.290$ & $+0.459$ & $+0.790$ & $+0.801$ & $+0.823$ & \cellcolor{ppocell}$\mathbf{+0.937}$ & $+0.914$ \\
Seoul, South Korea & 250 & $+0.555$ & $+0.604$ & $+0.973$ & $+0.930$ & $+0.980$ & \cellcolor{ppocell}$\mathbf{+0.994}$ & $+0.991$ \\
Sydney, Australia & 250 & $+0.090$ & $+0.147$ & $+0.589$ & $+0.475$ & $+0.520$ & \cellcolor{ppocell}$\mathbf{+0.702}$ & $+0.670$ \\
Tel Aviv, Israel & 75 & $+0.775$ & $+0.682$ & $+0.983$ & $+0.951$ & $+0.893$ & \cellcolor{ppocell}$\mathbf{+0.997}$ & \cellcolor{ppocell}$\mathbf{+0.997}$ \\
Tokyo, Japan & 75 & $+0.594$ & $+0.607$ & \cellcolor{bestcell}$\mathbf{+0.864}$ & $+0.756$ & $+0.699$ & $+0.810$ & $+0.845$ \\
Toronto, Canada & 250 & $+0.002$ & $+0.025$ & \cellcolor{bestcell}$\mathbf{+0.447}$ & $-0.223$ & $+0.407$ & $+0.368$ & $+0.383$ \\
\bottomrule
\end{tabular}
\end{table*}

On the base benchmark the learned controllers beat the SA-tuned rule, but that lead is
regime-bound.  For example, in the denser density-25 regime shown in
Table~\ref{tab:offdist_transfer}, every controller deployed unchanged falls behind SA in Cairo,
Paris, Tokyo and Toronto, where the tuned threshold rule becomes the best reactive controller.
Their advantage thus holds on the base scenario but not in every off-distribution regime.

\section{Conclusions}
\label{sec:conclusions}

We have shown that effective charging coordination for electric delivery fleets needs
no reservations, no messaging and no central dispatcher.  A compact controller shared
across the fleet, reading only its own state and a per-station public occupancy signal,
coordinates solely by reacting to those signals.  Several learners realize this local
interface, but not equally: PPO-joint is the strongest, coming within a mean reward gap
of $0.019$ of the omniscient Oracle across twenty international cities, transferring
zero-shot to the sixteen held out from training and holding up under contention and out of
distribution, while neuroevolution and a tuned threshold rule follow at a lower level.  NEAT
serves as a fast gradient-free search that validates that the interface is viable, but the greater
expressive capacity of PPO is essential for robustness out of distribution.  The
advantage of reading occupancy live shows up exactly in the congested cities where fixed
thresholds fail.  Another interesting property of PPO-joint is that it reaches this service
while charging about half as often as the Oracle.  The
controllers also rediscover strategies not encoded in the reward, such as partial charging and
routing around busy stations.

The paradigm is also practical to deploy.  Its one shared input, per-connector charger
status, is already produced by chargers over OCPP~\citep{OCA2020ocpp201} and can reach the
fleet through a standard MQTT publish--subscribe broadcast~\citep{Banks2019mqtt}, so it needs no new
infrastructure.  A decision is a compact-network forward pass of a few microseconds on a CPU,
with no GPU.

Several limitations remain.  The congestion model assumes stationary arrivals within a shift, and
charging costs and battery degradation are not modeled.  A natural next step is online
adaptation, letting the deployed controller keep learning as a city's conditions drift.  A
further extension is to model driver rest breaks.  They are unregulated in urban delivery, unlike
in long-haul transport, but a controller could still overlap charging with a break the shift
already spends, hiding much of its time cost.

\section*{CRediT authorship contribution statement}
\textbf{Javier Vales-Alonso:} Conceptualization, Methodology, Software, Validation, Formal analysis, Investigation, Writing -- Original Draft, Writing -- Review \& Editing, Visualization.
\textbf{Juan J. Alcaraz:}
Conceptualization, Methodology, Writing -- Review \& Editing.

\section*{Declaration of competing interest}
The authors declare that there are no known competing financial interests or personal relationships that could have appeared to influence the work reported in this paper.

\section*{Funding}
This work was supported by Google Cloud through an education credits grant.

\section*{Data availability}
The simulation code and the 20 frozen benchmark scenarios will be released in a public repository upon publication.

\section*{Declaration of generative AI and AI-assisted technologies in the manuscript preparation process}
During the preparation of this work the authors used a large language model (Claude, Anthropic) to assist in drafting and editing text and in preparing figures and tables from the authors' own data and results.  After using this tool, the authors reviewed and edited all content and take full responsibility for the content of the published article.

\bibliographystyle{elsarticle-harv}
\bibliography{references}

\end{document}